\documentclass{article}

\usepackage{microtype}
\usepackage{graphicx}
\usepackage{subcaption}
\usepackage{booktabs} 
\usepackage{amsmath, amsfonts, amssymb}
\usepackage{mathtools}
\usepackage{ifthen}
\usepackage{bm,bbm}
\graphicspath{{./figs/}}
\usepackage{hyperref}

\usepackage[accepted]{icml2019}

\icmltitlerunning{Quality of Uncertainty Quantification for Bayesian Neural Network Inference}

\begin{document}

\twocolumn[
  \icmltitle{Quality of Uncertainty Quantification for Bayesian Neural Network Inference}



\icmlsetsymbol{equal}{*}

\begin{icmlauthorlist}
\icmlauthor{Jiayu Yao}{equal,harvard}
\icmlauthor{Weiwei Pan}{equal,harvard}
\icmlauthor{Soumya Ghosh}{ibm}
\icmlauthor{Finale Doshi-Velez}{harvard}
\end{icmlauthorlist}

\icmlaffiliation{harvard}{Harvard University, Cambridge, MA}
\icmlaffiliation{ibm}{IBM T.J. Watson Research Center, Cambridge, MA}

\icmlcorrespondingauthor{Jiayu Yao}{jiy328@g.harvard.edu}

\icmlkeywords{}

\vskip 0.3in
]



\printAffiliationsAndNotice{\icmlEqualContribution} 

\begin{abstract}
Bayesian Neural Networks (BNNs) place priors over the parameters in a neural network. Inference in BNNs, however, is difficult; all inference methods for BNNs are approximate. In this work, we empirically compare the quality of predictive uncertainty estimates for 10 common inference methods on both regression and classification tasks. Our experiments demonstrate that commonly used metrics (e.g. test log-likelihood) can be misleading. Our experiments also indicate that inference innovations designed to capture structure in the posterior do not necessarily produce high quality posterior approximations. 
\end{abstract}


\section{Introduction}\label{sec:intro}
While deep learning provides a flexible framework for function approximation that achieves impressive performance on many real-life tasks~\citep{lecun2015deep}, there has been a recent focus on providing predictive uncertainty estimates for deep models, making them better suited for use in risk-sensitive applications. Bayesian neural networks (BNNs) are neural network models that include uncertainty through priors on network weights, and thus provide  uncertainty about the functional mean through posterior predictive distributions~\citep{mackay1992practical,neal2012bayesian}.  (Note: one can also place priors directly on functions rather than network weights~\citep{sun2019functional}; in this work, we focus on the more commonly used approach of placing priors over weights.) 

Unfortunately, characterizing uncertainty over parameters of neural networks is challenging due to the high-dimensionality of the weight space and potentially complex dependencies among the weights. Markov-chain Monte Carlo (MCMC) techniques are often slow to mix. Standard variational inference methods with mean field approximations may struggle to escape local optima and furthermore, are unable to capture dependencies between the weights.

There exists a large body of work to improve the quality of inference for Bayesian neural networks (BNNs) by improving the approximate inference procedure (e.g. ~\citealt{graves2011practical, blundell2015weight, hernandez2016black}, to name a few), or by improving the flexibility of the variational approximation for variational inference (e.g. \citealt{gershman2012nonparametric,ranganath2016hierarchical, louizos2017multiplicative, miller2017variational}). On the other hand, a number of frequentist approaches, like ensemble methods \cite{lakshminarayanan2017simple,pearce2018uncertainty,tagasovska2018frequentist}, provide predictive uncertainty estimates for neural network while by-passing the challenges of Bayesian inference all together.

The objective of this work is to provide an empirical comparison of common BNN inference approaches with a focus on the quality of uncertainty quantification.  We perform a careful empirical comparison of 8 state-of-the-art approximate inference methods and 2 non-Bayesian frameworks, where we find that performance depends heavily on the training data. We characterize situations where metrics like log-likelihood and RMSE fail to distinguish good vs poor approximations of the true posterior, and, based on our observations, engineer synthetic datasets for comparing the predictive uncertainty estimates.


\section{ Related Works}\label{sec:lit}
In literature, posteriors for Bayesian Neural Network models obtained by Hamiltonian Monte Carlo (HMC) \cite{neal2012bayesian}
are frequently used as ground truth. However, HMC scales poorly on high dimensional parameter space and large datasets \cite{welling2011bayesian, chen2014stochastic}. Mini-batched versions of HMC, such as Stochastic Gradient Langevin Dynamics (SGLD) \cite{welling2011bayesian} and Stochastic Gadient HMC \cite{chen2014stochastic}, have been introduced to address the issue of scalability. However, these methods still suffer from lower mixing rate and are not theoretically guaranteed to converge to the true posterior when model assumptions are not met (e.g. when the true model of the gradient noise is not well-estimated).

As a result, much effort has been spent on variational methods.  Mean Field Variational Bayes for BNNs were in introduced in \cite{graves2011practical}, the gradient computation of which was later improved in Bayes by Backprop (BBB)~\citep{blundell2015weight}. However, the fully factorized Gaussian variational family used in BBB is unable to capture correlation amongst the parameters in the posterior. In contrast, Matrix Gaussian Posteriors (MVG)~\cite{louizos2016structured}, Multiplicative Normalizing Flows (MNF)~\cite{louizos2017multiplicative}, and Bayes by Hypernet (BBH)~\cite{2017arXiv171101297P} are explicitly designed to capture posterior correlation by imposing structured approximation families; works like Black Box $\alpha$-Divergence \cite{hernandez2016black} and Probabilistic Backpropagation (PBP) \citep{hernandez2015probabilistic} use a richer family of divergence measures, encouraging approximate posteriors to capture important properties of true posterior distributions. 

Finally, Dropout \cite{gal2016dropout} and ensemble methods \cite{lakshminarayanan2017simple,pearce2018uncertainty,tagasovska2018frequentist} by-pass the difficulties of performing Bayesian inference and obtain predictive uncertainty estimates through implicitly or explicitly training multiple models on the same data.

While there are numerous inference methods, there have been few exhaustive, independent comparisons \cite{myshkov2016posterior, zhaoempirical}. In \cite{myshkov2016posterior} BBB, PBP, and Dropout are compared with mini-batched HMC on regression. The evaluation metrics consist of RMSE and divergence from HMC posteriors (considered as ground truth). In \cite{zhaoempirical}, BBB and Dropout are compared with HMC and SGHMC on classification. Accuracy and calibration (how well predictive uncertainty align with empirical uncertainty) of the posterior predictive distribution are analyzed. Neither work indicates when predictive accuracy and calibration correspond to the fidelity of posterior approximation.

In this work, we provide a comparison of a wide range of inference methods on both regression and classification tasks. Furthermore, we investigate the usefulness of metrics for posterior predictive generalization and calibration for measuring the fidelity of posterior approximations. In particular, we identify situations in which these metrics are poor proxies for measuring divergence from true posteriors.

\section{Background}\label{sec:back}

Let $\mathcal{D}=\{(x_1,y_1), \ldots (x_N, y_N) \}$ be a dataset of $N$ observations.  Each input $x_n \in \mathbb{R}^D$ is a $D$-dimensional vector and each output $y_n\in \mathbb{R}^K$ is $K$-dimensional. In classification problems, $y_n$ is the vector of probabilities of $K$-classes.

A \emph{Bayesian Neural Network (BNN)} assumes a likelihood of the form $y = f(x; W) + \epsilon$, where $f$ is a neural network parametrized by $W$ and $\epsilon$ is a normally distributed noise variable.  Given a prior over weights $p(W)$, uncertainty in a BNN is modeled by a posterior, $p(W| \mathcal{D})$. At test time, predictions are made via the posterior predictive distribution, $p(y | x, \mathcal{D})$:
\begin{align}
p(y| x, \mathcal{D}) = \int p(y| x, W) p(W| \mathcal{D}) dW. 
\end{align}

\section{Challenges in Evaluating Uncertainty}
\begin{figure*}[t]
    \centering
        \includegraphics[width=0.23\textwidth]{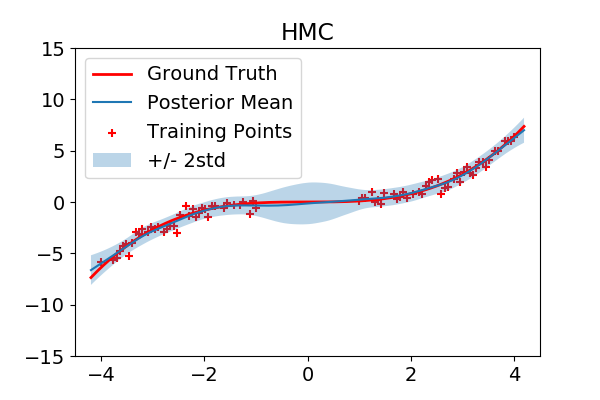} 
    ~ 
        \includegraphics[width=0.23\textwidth]{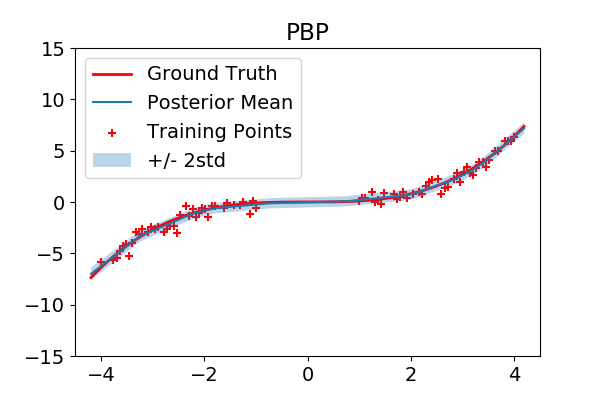} 
    ~
        \includegraphics[width=0.23\textwidth]{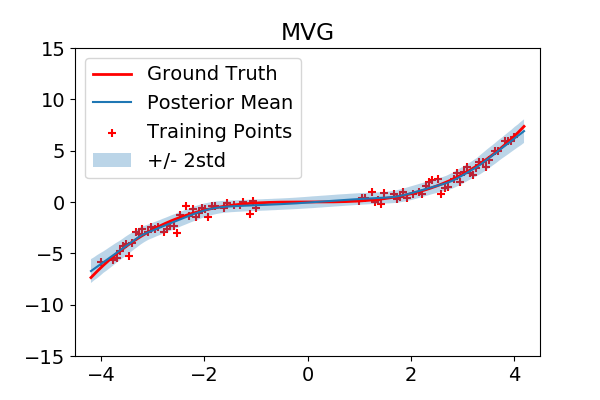}
    ~ 
        \includegraphics[width=0.23\textwidth]{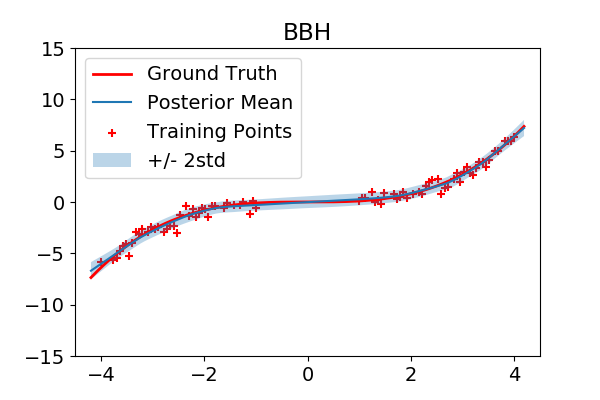}
    \caption{A comparison of the posterior predictives. Ground truth (HMC) reveals that our BNN model class perhaps has more flexibility than needed (as indicated by the widening in the predictive posterior where there are no data). BBB, MVG and BBH produce approximate posterior predictives that incorrectly have lower variance but all have test log-likelihoods that are comparable if not higher to that of the ground truth. More comparisons in Appendix \ref{sec:results}.}
     \label{fig:pedagogical}
\end{figure*}

Frequently in literature, high test log likelihood is used as evidence that the inference procedure has more faithfully captured the true posterior.  However, here we argue that while test log likelihood may be a good criteria for model selection, it is not a reliable criteria for determining how well an approximate posterior aligns with the true posterior.  

Consider the example in Figure \ref{fig:pedagogical}. The training data has a `gap', namely there are no samples from $[-1, 1]$.  We see that the posterior predictive means of the true posterior (i.e. the ground truth), as given by HMC (details in Section \ref{sec:exp}), and that obtained by PBP are identical.  However, the PBP posterior predictive uncertainty is far smaller. The average test log-likelihood for data evenly spaced in $[-4,4]$ is -0.25 for PBP and -0.42 for HMC.  In this case, the better number does not indicate a better model class (e.g. a prior $p(W)$ that appropriately puts more weight where the data lie). Rather, it is an artifact of the fact that the data happens to lie where an incorrect inference procedure put more mass. In short, the average test log-likelihood indicates that the approximate posterior predictive better aligns with the data and not that it is a faithful approximation of the true posterior predictive.

For the same reason, RMSE and other metrics for measuring predictive calibration (such as Prediction Interval Coverage Probability) are also unreliable indicators of the degree to which approximate posteriors align with the true ones. In this paper, we argue that issues of model selection should be addressed separately from issues associated with the approximation gaps of inference. For this, we engineer synthetic datasets on which our ground-truth BNN model produces well-calibrated posterior predictive distributions and hence generalization and calibration metrics are proxies for how well a given inference method captures the true posterior.


\section{Experimental Set Up}\label{sec:exp}

\paragraph{Data Sets}
We perform experiments on univariate regression and two-dimensional binary classification tasks so that the ground truth distributions can be visualized. For each task, we consider two synthetic datasets. In one of these datasets, the a priori model uncertainty will be  higher than the variation in the data warrants, whereas in the other dataset the data variation will match the a priori model uncertainty. Data generation details are in Appendix \ref{sec:data}.

\paragraph{Ground Truth Baselines}
We use Hamiltonian Monte Carlo (HMC) \cite{neal2012bayesian} to construct `ground-truth' posterior and posterior predictive distributions. We run HMC for 50k iterations with 100 leapfrog steps and check for mixing. See Appendix \ref{hyperparameter} for full description.

\paragraph{Methods}
We evaluate 10 inference methods: Bayes by Backprop (BBB), Probabilistic Backpropagation (PBP), Black-box $\alpha$-Divergence (BB-ALPHA), Multiplicative Normalizing Flows (MNF), Matrix-Variate Gaussian (MVG), Bayes by Hypernet (BBH), Dropout, Ensemble, Stochastic Gradient Langevin  Dynamics (SGLD), and Stochastic Gradient HMC (SGHMC).
We do not evaluate PBP and BB-ALPHA on classification tasks as they assume exponential family as likelihood distributions.
All optimization is done with Adam except for HMC, SGLD and SG-HMC which have their own scheduled gradient updates. We use existing code-bases for methods when available (BB-ALPHA, MVG, BBH, NMF). Full description of tuning schemes is in Appendix \ref{hyperparameter}.

\paragraph{Experimental Parameters}
For all tasks, we use neural networks with ReLU nonlinearities. We use 1 hidden layer with 50 hidden nodes for regression and 2 hidden layers with 10 nodes each for classification. Every method is run with 20 random restarts, each until convergence, using a fixed weight prior $W\sim \mathcal{N}(0, I)$ and true output noise. For methods that include priors on the output noise, we disable these in the experiments. Out of the 20 restarts, we select the solution with the highest validation log-likelihood and estimated the posterior predictive distributions with 500 posterior samples (results given by selection by ELBO are indistinguishable and are in Appendix \ref{sec:results}). Full description in Appendix \ref{hyperparameter}.

\paragraph{Evaluation Metrics}
Evaluating the fidelity of posterior approximations is challenging.  
As a result, in BNN literature, accuracy, average marginal log-likelihood, and frequentist metrics such as Prediction Interval Coverage Probability (PICP) -- the percentage of observations for which the ground truth $y$ lies within a $95\%$ predictive-interval (PI) of the learned model -- and the average width of the $95\%$ PI (MPIW) are commonly used as indicators of the quality of posterior approximation (full description of metrics in Appendix \ref{sec:metrics}). 
Our experiments provide insights on when these metrics correspond with high quality posterior approximation and when they do not.

\section{Results}\label{sec:results}

\begin{figure*}[h!]
    \centering
     \begin{subfigure}[t]{0.48\textwidth}
        \centering
        \includegraphics[width=\textwidth]{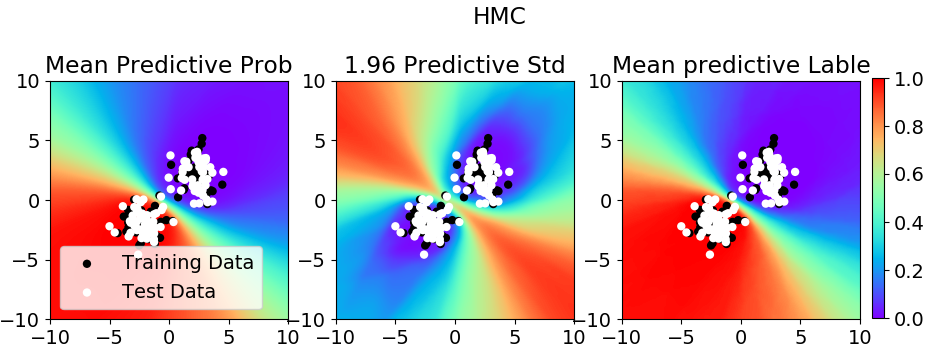} 
        \caption{\footnotesize{HMC posterior predictive mean over probabilities, posterior predictive standard deviation and posterior predctive mean over labels (from left to right)}}
    \end{subfigure}%
    ~ 
    \begin{subfigure}[t]{0.48\textwidth}
        \centering
        \includegraphics[width=\textwidth]{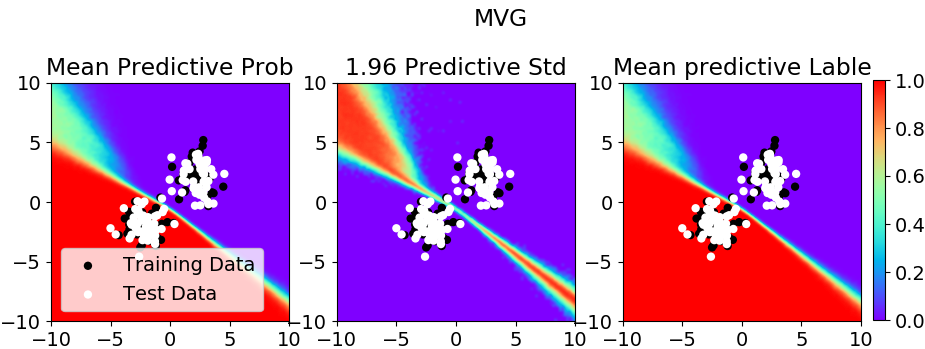} 
        \caption{\footnotesize{MVG posterior predictive mean over labels, posterior predictive standard deviation and posterior predctive mean over probabilities (from left to right)}}
    \end{subfigure}
    \caption{\footnotesize{Ground truth (HMC) indicates that the a priori model uncertainty is overly high. MVG produces approximate posterior predictives that have lower uncertainty than the ground truth but have test log-likelihoods and ROC's that are identical to the ground truth. More comparisons in Appendix \ref{sec:results}.}}
    \label{fig:pedagogical2}
\end{figure*}

\begin{figure*}[h!]
    \centering
        \includegraphics[width=0.18\textwidth]{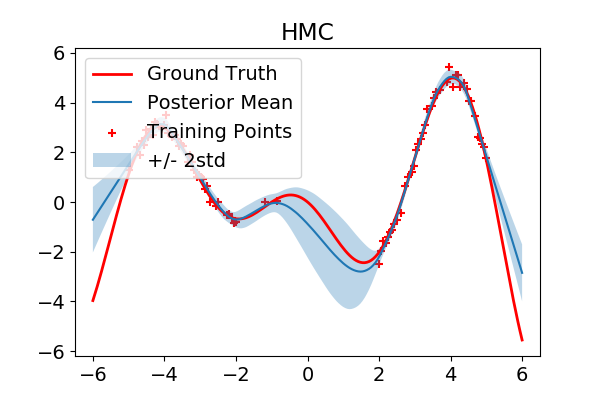} 
    ~
         \includegraphics[width=0.18\textwidth]{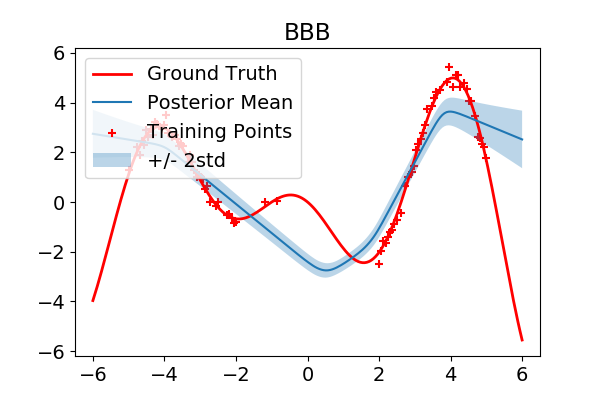} 
   ~ 
         \includegraphics[width=0.18\textwidth]{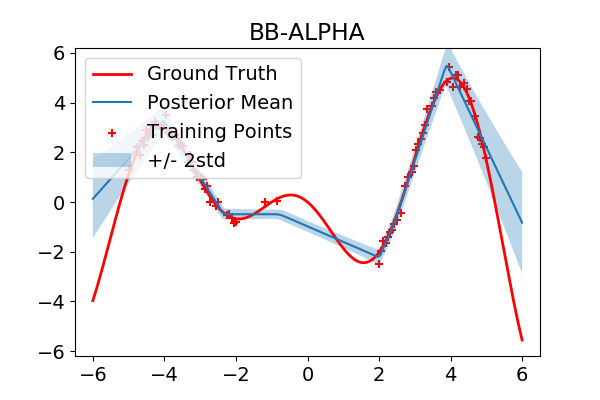} 
    ~ 
         \includegraphics[width=0.18\textwidth]{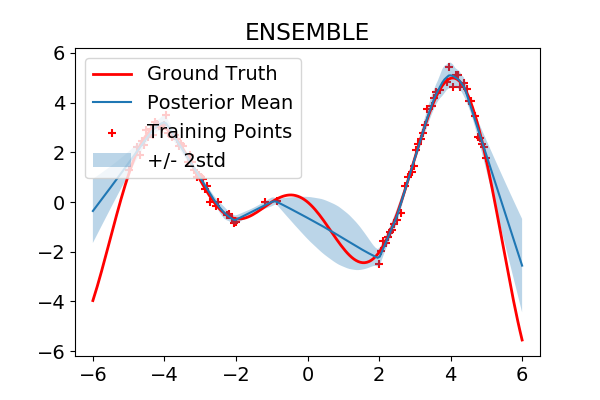}
         ~ 
         \includegraphics[width=0.18\textwidth]{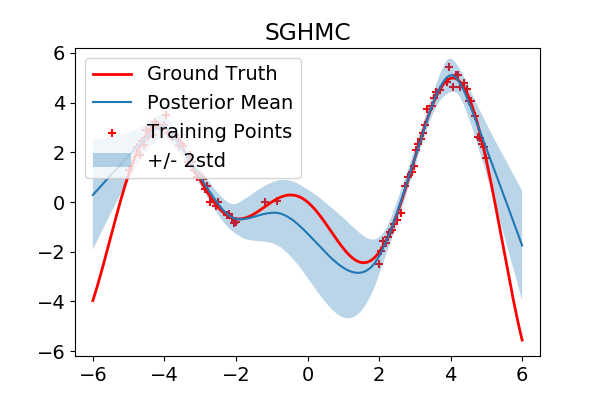}

    \caption{A comparison of the posterior predictives for Regression with Matched A Priori Uncertainty. Ground truth (HMC) indicates that the model is calibrated. All inference methods produce posterior predictives that model the data well over regions well represented in the training data. Of these methods, all, except for Ensemble and SGHMC, underestimate uncertainty over regions sparsely represented in the training data. More comparisons in Appendix \ref{sec:results}.}
    \label{fig:reg2}
\end{figure*}

\setlength{\tabcolsep}{3.0pt}
\begin{table*}[h!]
\scriptsize
\centering
\begin{tabular}{c||c|c|c|c|c|c|c|c|c|c|c}
&\emph{HMC}&\emph{BBB}&\emph{PBP}&\emph{BB-$\alpha$}&\emph{MVG}&\emph{MNF}&\emph{BbH}&\emph{Dropout}&\emph{Ensemble}&\emph{SGLD}&\emph{SGHMC}\\
\hline
\hline
RMSE&\textbf{0.85}$\pm$0.01&2.33$\pm$0.11&2.92$\pm$0.29&1.86$\pm$1.65&1.70$\pm$0.37&\textbf{1.11}$\pm$0.21&1.32$\pm$0.22&1.45$\pm$0.17&0.90$\pm$0.01&0.86$\pm$0.08&1.18$\pm$0.01\\ 
\hline
LogLL&\textbf{-1.40}$\pm$0.28&-41.12$\pm$6.23&-106.78$\pm$19.64&-5.41$\pm$2.82&-26.69$\pm$12.18&-13.85$\pm$6.87&-12.55$\pm$8.23&-5.99$\pm$1.82&-6.65$\pm$0.09&-3.60$\pm$0.75&\textbf{-1.27}$\pm$0.26\\ 
\hline
PICP&\textbf{0.86}$\pm$0.00&0.46$\pm$0.04&0.32$\pm$0.07&0.78$\pm$0.09&0.64$\pm$0.03&0.63$\pm$0.04&0.65$\pm$0.03&0.60$\pm$0.04&0.84$\pm$0.00&0.75$\pm$0.02&\textbf{0.86}$\pm$0.00\\ 
\hline
MWPI&1.79$\pm$0.02&1.67$\pm$0.04&0.81$\pm$0.00&6.57$\pm$12.74&1.47$\pm$0.20&0.92$\pm$0.02&1.21$\pm$0.19&1.64$\pm$0.18&1.50$\pm$0.05&1.37$\pm$0.08&2.62$\pm$0.04\\ 
\end{tabular}
\caption{A comparison of generalization and calibration metrics for Regression with Matched A Priori Uncertainty. 
Unlike in the first regression task (Figure \ref{fig:pedagogical}), generalization metrics combined with calibration metrics give a reasonable indication for the quality of posterior approximation (HMC scores highest). However, even here these metric do not entirely capture our intuition for quality of fit (for example, the test log-likelihood of BB-ALPHA is higher than Ensemble).}
 \label{tab:reg2}
\end{table*}
\normalsize

\paragraph{Generalization and calibration metrics are not reliable indicators for quality of posterior approximation.}
Figure \ref{fig:pedagogical} and \ref{fig:pedagogical2} show that when the model class has large flexibility for the data, the ground truth posterior predictive may have lower log-likelihood and calibration scores than a poor approximation. In Figure \ref{fig:pedagogical}, we see most inference methods, though underestimate the uncertainty, still produce high log-likelyhood because the predictive mean aligns well with the true function. But HMC gets penalized by giving large uncertainty in the middle due to model class flexibility. On the other hand, when the model class has the right capacity for the data, posterior predictive generalization and calibration are good but not definitive indicators of the quality of posterior approximation (Figure \ref{fig:reg2}, Table \ref{tab:reg2}). This is especially concerning for high-dimensional or large datasets on which ground truth distributions are hard to compute and appropriate model capacity is hard to ascertain. Here, generalization/calibration metrics often conflate bad models with bad inference. We note that evaluations of uncertainty estimates based on active learning will struggle similarly in distinguishing model and inference issues.

\paragraph{Inference methods designed to capture structure in the posterior do not necessarily produce better approximations of the true posterior.} In our experiments, we do not see that methods using a richer divergence metric or structured variational family are able to better capture the ground truth posterior. This is likely due to the fact that the true posteriors in our experiments lack patterns of dependencies that those inference methods aim to capture (Appendix \ref{sec:posterior}).
However, this observation indicates the need for developing concrete guidelines for when it is beneficial to use alternative divergence metrics and structured variational families, since the extra flexibility of these methods often invites additional optimization challenges on real-data. 

\paragraph{Ensemble methods do not consistently produce the types of uncertainty estimates we want.} Methods using an ensemble (whether explicit or implicit) of models to produce predictive distributions rely on the model diversity to produce accurate uncertainty estimates. Ensemble methods may produce similar solutions due to initialization or optimization issues. When the ensemble includes many dissimilar plausible models for the data (Figure \ref{fig:reg2}) the uncertainty estimate can be good; when ensemble training finds local optima with highly similar models for the data the uncertain estimates can be poor (Figure \ref{fig:class2}). Thus, uncertainty estimates from ensembles can be unreliable absent a structured way of including diversity training objectives. 

\paragraph{SGHMC produces posterior predictives that are most similar to that of HMC.} In our experiments, we see that SGLD  
drastically underestimates posterior predictive uncertainty. SGHMC, while tending to overestimate uncertainty, produces predictive distributions qualitatively similar to those of HMC.

\section{Conclusion}\label{sec:conclude}
In this paper, we compare 10 commonly used approximate inference procedures for Bayesian Neural Networks. Frequently, measurements of generalization and calibration of the posterior predictive are used to evaluate the quality of inference. We show that these metrics conflate issues of model selection with those of inference. On our data, we see that approximate Bayesian inference methods struggle to capture true posteriors and the non-Bayesian methods often do not capture the type of predictive uncertainty that we want. Our experiments show that we need more exhaustive and standardized evaluation of new, complex approximate inference methods. Furthermore, we need careful metrics for formalizing desiderata we have for uncertainty estimation.  

\textbf{Acknowledgements}
JY acknowledges support from NSF RI-1718306. SG was supported by an IBM Faculty Research award. WP was supported by the Institute for Applied Computational Science. FD was supported by AFOSR FA 9550-17-1-0155.

\bibliography{bib}

\begin{thebibliography}{22}
\providecommand{\natexlab}[1]{#1}
\providecommand{\url}[1]{\texttt{#1}}
\expandafter\ifx\csname urlstyle\endcsname\relax
  \providecommand{\doi}[1]{doi: #1}\else
  \providecommand{\doi}{doi: \begingroup \urlstyle{rm}\Url}\fi

\bibitem[Blundell et~al.(2015)Blundell, Cornebise, Kavukcuoglu, and
  Wierstra]{blundell2015weight}
Blundell, C., Cornebise, J., Kavukcuoglu, K., and Wierstra, D.
\newblock Weight uncertainty in neural networks.
\newblock \emph{arXiv preprint arXiv:1505.05424}, 2015.

\bibitem[Chen et~al.(2014)Chen, Fox, and Guestrin]{chen2014stochastic}
Chen, T., Fox, E., and Guestrin, C.
\newblock Stochastic gradient hamiltonian monte carlo.
\newblock In \emph{International Conference on Machine Learning}, pp.\
  1683--1691, 2014.

\bibitem[Gal \& Ghahramani(2016)Gal and Ghahramani]{gal2016dropout}
Gal, Y. and Ghahramani, Z.
\newblock Dropout as a bayesian approximation: Representing model uncertainty
  in deep learning.
\newblock In \emph{international conference on machine learning}, pp.\
  1050--1059, 2016.

\bibitem[Gershman et~al.(2012)Gershman, Hoffman, and
  Blei]{gershman2012nonparametric}
Gershman, S., Hoffman, M., and Blei, D.
\newblock Nonparametric variational inference.
\newblock \emph{arXiv preprint arXiv:1206.4665}, 2012.

\bibitem[Graves(2011)]{graves2011practical}
Graves, A.
\newblock Practical variational inference for neural networks.
\newblock In \emph{Advances in neural information processing systems}, pp.\
  2348--2356, 2011.

\bibitem[Hern{\'a}ndez-Lobato \& Adams(2015)Hern{\'a}ndez-Lobato and
  Adams]{hernandez2015probabilistic}
Hern{\'a}ndez-Lobato, J.~M. and Adams, R.
\newblock Probabilistic backpropagation for scalable learning of bayesian
  neural networks.
\newblock In \emph{International Conference on Machine Learning}, pp.\
  1861--1869, 2015.

\bibitem[Hern{\'a}ndez-Lobato et~al.(2016)Hern{\'a}ndez-Lobato, Li, Rowland,
  Hern{\'a}ndez-Lobato, Bui, and Turner]{hernandez2016black}
Hern{\'a}ndez-Lobato, J.~M., Li, Y., Rowland, M., Hern{\'a}ndez-Lobato, D.,
  Bui, T., and Turner, R.~E.
\newblock Black-box $\alpha$-divergence minimization.
\newblock 2016.

\bibitem[Lakshminarayanan et~al.(2017)Lakshminarayanan, Pritzel, and
  Blundell]{lakshminarayanan2017simple}
Lakshminarayanan, B., Pritzel, A., and Blundell, C.
\newblock Simple and scalable predictive uncertainty estimation using deep
  ensembles.
\newblock In \emph{Advances in Neural Information Processing Systems}, pp.\
  6402--6413, 2017.

\bibitem[LeCun et~al.(2015)LeCun, Bengio, and Hinton]{lecun2015deep}
LeCun, Y., Bengio, Y., and Hinton, G.
\newblock Deep learning.
\newblock \emph{nature}, 521\penalty0 (7553):\penalty0 436, 2015.

\bibitem[Louizos \& Welling(2016)Louizos and Welling]{louizos2016structured}
Louizos, C. and Welling, M.
\newblock Structured and efficient variational deep learning with matrix
  gaussian posteriors.
\newblock In \emph{International Conference on Machine Learning}, pp.\
  1708--1716, 2016.

\bibitem[Louizos \& Welling(2017)Louizos and
  Welling]{louizos2017multiplicative}
Louizos, C. and Welling, M.
\newblock Multiplicative normalizing flows for variational bayesian neural
  networks.
\newblock \emph{arXiv preprint arXiv:1703.01961}, 2017.

\bibitem[MacKay(1992)]{mackay1992practical}
MacKay, D.~J.
\newblock A practical bayesian framework for backpropagation networks.
\newblock \emph{Neural computation}, 4\penalty0 (3):\penalty0 448--472, 1992.

\bibitem[Miller et~al.(2017)Miller, Foti, and Adams]{miller2017variational}
Miller, A.~C., Foti, N.~J., and Adams, R.~P.
\newblock Variational boosting: Iteratively refining posterior approximations.
\newblock In \emph{Proceedings of the 34th International Conference on Machine
  Learning-Volume 70}, pp.\  2420--2429. JMLR. org, 2017.

\bibitem[Myshkov \& Julier(2016)Myshkov and Julier]{myshkov2016posterior}
Myshkov, P. and Julier, S.
\newblock Posterior distribution analysis for bayesian inference in neural
  networks.
\newblock 2016.

\bibitem[Neal(2012)]{neal2012bayesian}
Neal, R.~M.
\newblock \emph{Bayesian learning for neural networks}, volume 118.
\newblock Springer Science \& Business Media, 2012.

\bibitem[{Pawlowski} et~al.(2017){Pawlowski}, {Brock}, {Lee}, {Rajchl}, and
  {Glocker}]{2017arXiv171101297P}
{Pawlowski}, N., {Brock}, A., {Lee}, M.~C.~H., {Rajchl}, M., and {Glocker}, B.
\newblock {Implicit Weight Uncertainty in Neural Networks}.
\newblock \emph{ArXiv e-prints}, November 2017.

\bibitem[Pearce et~al.(2018)Pearce, Zaki, Brintrup, and
  Neel]{pearce2018uncertainty}
Pearce, T., Zaki, M., Brintrup, A., and Neel, A.
\newblock Uncertainty in neural networks: Bayesian ensembling.
\newblock \emph{arXiv preprint arXiv:1810.05546}, 2018.

\bibitem[Ranganath et~al.(2016)Ranganath, Tran, and
  Blei]{ranganath2016hierarchical}
Ranganath, R., Tran, D., and Blei, D.
\newblock Hierarchical variational models.
\newblock In \emph{International Conference on Machine Learning}, pp.\
  324--333, 2016.

\bibitem[Sun et~al.(2019)Sun, Zhang, Shi, and Grosse]{sun2019functional}
Sun, S., Zhang, G., Shi, J., and Grosse, R.
\newblock Functional variational bayesian neural networks.
\newblock \emph{arXiv preprint arXiv:1903.05779}, 2019.

\bibitem[Tagasovska \& Lopez-Paz(2018)Tagasovska and
  Lopez-Paz]{tagasovska2018frequentist}
Tagasovska, N. and Lopez-Paz, D.
\newblock Frequentist uncertainty estimates for deep learning.
\newblock \emph{arXiv preprint arXiv:1811.00908}, 2018.

\bibitem[Welling \& Teh(2011)Welling and Teh]{welling2011bayesian}
Welling, M. and Teh, Y.~W.
\newblock Bayesian learning via stochastic gradient langevin dynamics.
\newblock In \emph{Proceedings of the 28th International Conference on Machine
  Learning (ICML-11)}, pp.\  681--688, 2011.

\bibitem[Zhao \& Ji(2018)Zhao and Ji]{zhaoempirical}
Zhao, R. and Ji, Q.
\newblock An empirical evaluation of bayesian inference methods for bayesian
  neural networks.
\newblock 2018.

\end{thebibliography}
\bibliographystyle{icml2019}

 

\appendix

\section{Data} \label{sec:data}
Regression taks:
\begin{enumerate}
\item \textbf{Regression with Mismatched A Priori Uncertainty:} targets are given by $y = 0.1x^3+\epsilon$, where $\epsilon \sim N(0, 0.25)$. Evaluated on 80 training inputs, 20 validation inputs uniformly sampled from $[-4, -1] \cup [1, 4]$ and 200 test inputs uniformly sampled from $[-4, 4]$. 
\item \textbf{Regression with Matched A Priori Uncertainty:} targets are given by $y =-(1+x)sin(1.2x)+\epsilon$, where $\epsilon \sim N(0, 0.04)$. Evaluated on 80 training inputs, 20 validation inputs uniformly sampled from $[-6, -2] \cup [2, 6]$, 2 training inputs, 2 validation inputs uniformly sampled from $[-2, 2]$, and 200 test inputs uniformly sampled from $[-6, 6]$.
\end{enumerate}
Classification taks:
\begin{enumerate}
\item \textbf{Classification with Mismatched A Priori Uncertainty:} 2 balanced classes targets are generated from two multivariate Gaussian distribution $p_1\sim\mathcal{N}\bigg(\begin{bmatrix}2\\2\end{bmatrix},\mathbb{I}\bigg),p_2\sim\mathcal{N}\bigg(\begin{bmatrix}-2\\-2\end{bmatrix},\mathbb{I}\bigg)$. Evaluated on 80 training inputs, 20 validation inputs and 100 test inputs uniformly sampled from $p_1,\ p_2$.
\item \textbf{Classification with Matched A Priori Uncertainty:} 2 balanced classes targets are generated from two multivariate Gaussian distribution $p_1\sim\mathcal{N}\bigg(\begin{bmatrix}3\\0\end{bmatrix},\Sigma\bigg),p_2\sim\mathcal{N}\bigg(\begin{bmatrix}-3\\0\end{bmatrix},\Sigma\bigg)$ where $\Sigma=\begin{bmatrix}2&1\\1&2\end{bmatrix}$. Evaluated on 80 training inputs, 20 validation inputs sampled from $p_1,\ p_2$ with Class 1 truncated with $x^{(2)}\leq0$ and Class 2 truncated with $x^{(2)}\geq0$. 100 test inputs are uniformly sampled from $p_1,\ p_2$.
\end{enumerate}


\section{Evaluation Metrics} \label{sec:metrics}
The \emph{average marginal log-likelihood} is computed as: 
\begin{equation}
\mathbb{E}_{(x_n,y_n)\sim\mathcal{D}}\left\lbrack\mathbb{E}_{q(W)} \left\lbrack p(y_n | x_n, W) \right\rbrack\right\rbrack.
\label{eq:avg-ll}
\end{equation}
The \emph{predictive RMSE} is computed as:
\begin{equation}
\sqrt{\frac{1}{N} \sum\limits_{n=1}^N  \lVert y_n -  \mathbb{E}_{q(W)} \left[f(x_n, W)\right] \rVert_2^2}.
\label{eqn:mse}
\end{equation}
The Prediction Interval Coverage Probability (PICP) is computed as:
\begin{align}
\frac{1}{N} \sum_{n=1}^N \mathbbm{1}_{y_n\leq\widehat{y}^{high}_n}\cdot\mathbbm{1}_{y_n\geq\widehat{y}^{low}_n},
\end{align}
and the Mean Prediction Interval Width (MPIW) is computed as:
\begin{align}
\frac{1}{N} \sum_{n=1}^N \left(\widehat{y}^{high}_n - \widehat{y}^{low}_n\right),
\end{align}
where $\widehat{y}^{high}_n$ is the $97.5\%$ percentile and $\widehat{y}^{low}_n$ is the $2.5\%$ percentile of the predicted outputs for $x_n$. We want models to have PICP values to be close to $95\%$ while minimizing the MPIW, thus formalizing our desiderata that well-calibrated posterior predictive uncertainty should be both necessary and sufficient to capture the variation in the data.

\section{Additional Results}\label{sec:more_results}
\begin{figure*}[h!]
    \centering

        \includegraphics[width=0.18\textwidth]{toy_reg1_hmc.png} 
    ~
         \includegraphics[width=0.18\textwidth]{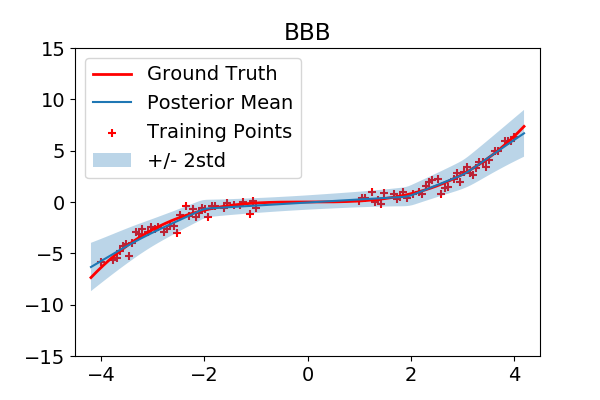} 
   ~ 
        \includegraphics[width=0.18\textwidth]{toy_reg1_pbp.png} 
    ~ 
        \includegraphics[width=0.18\textwidth]{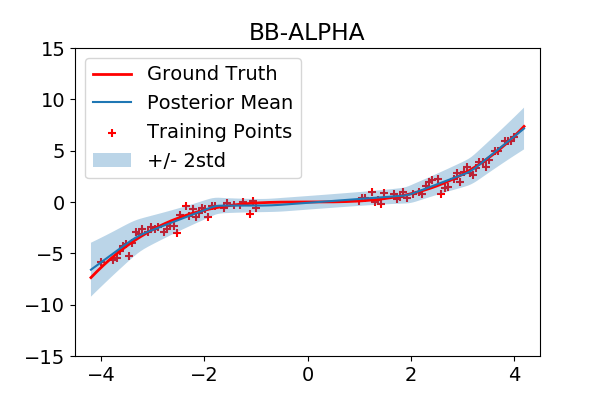} 
    ~ 
         \includegraphics[width=0.18\textwidth]{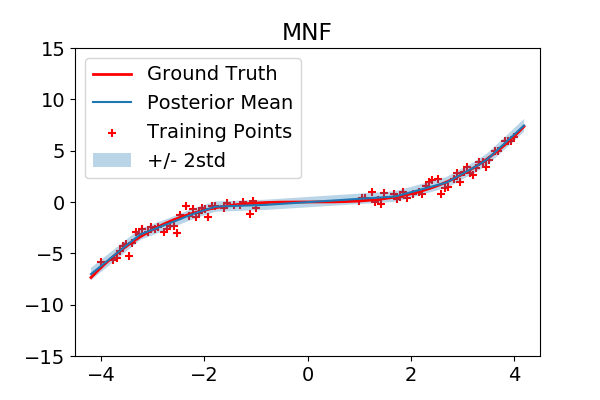} 
   ~ 
         \includegraphics[width=0.18\textwidth]{toy_reg1_mvg.png}
    ~
         \includegraphics[width=0.18\textwidth]{toy_reg1_bbh.png} 
   ~ 
         \includegraphics[width=0.18\textwidth]{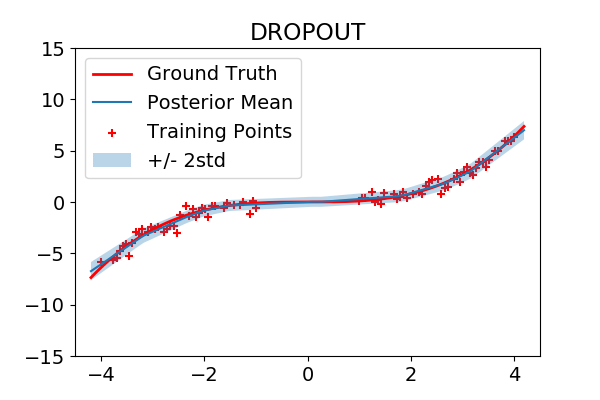}
    ~ 
         \includegraphics[width=0.18\textwidth]{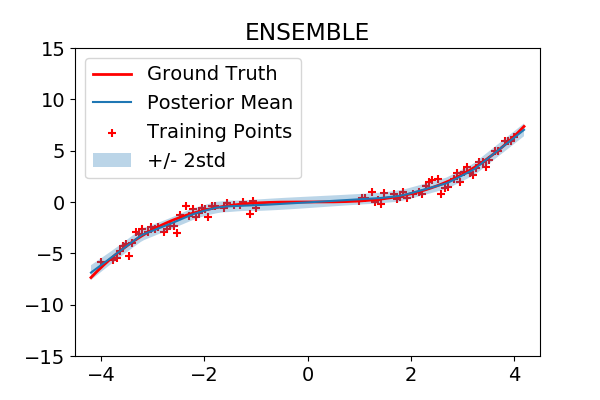}
    ~ 
         \includegraphics[width=0.18\textwidth]{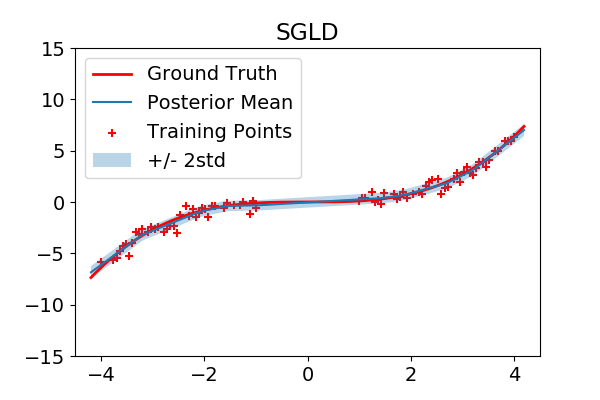}
         ~ 
         \includegraphics[width=0.18\textwidth]{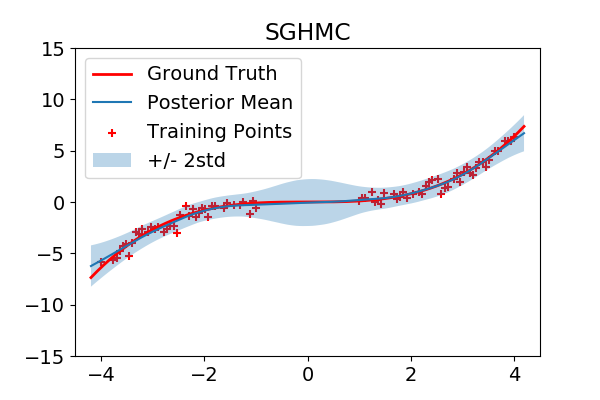}

    \caption{A comparison of the posterior predictives for Regression with Mimatched A Priori Uncertainty. Ground truth (HMC) indicates that the prior over estimates the variations in the data. All methods, except for SGHMC, produce approximate posterior predictives that have lower variance but all have test log-likelihoods that are comparable if not higher to that of the ground truth.}
    \label{fig:reg1}
\end{figure*}

\begin{figure*}[h!]
    \centering
        \includegraphics[width=0.18\textwidth]{toy_reg2_hmc.png} 
    ~
         \includegraphics[width=0.18\textwidth]{toy_reg2_bbb.png} 
   ~ 
        \includegraphics[width=0.18\textwidth]{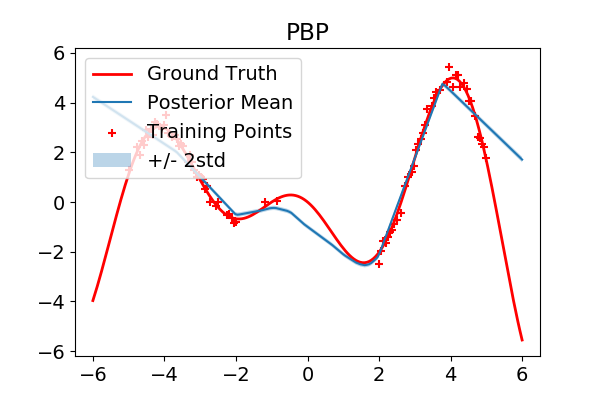} 
    ~ 
        \includegraphics[width=0.18\textwidth]{toy_reg2_bb-alpha.png} 
    ~ 
         \includegraphics[width=0.18\textwidth]{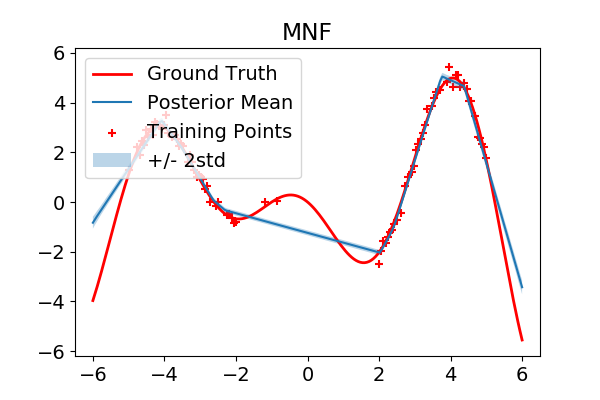} 
   ~ 
         \includegraphics[width=0.18\textwidth]{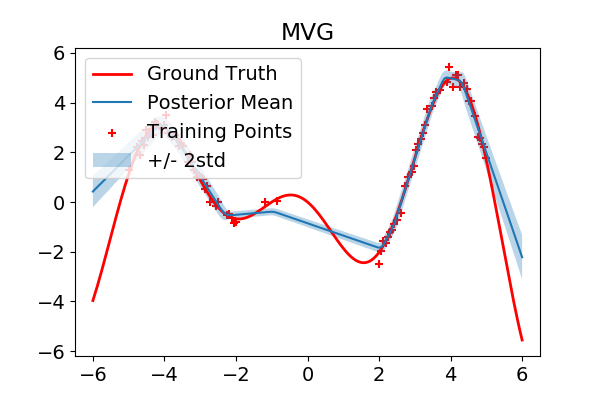}
    ~
         \includegraphics[width=0.18\textwidth]{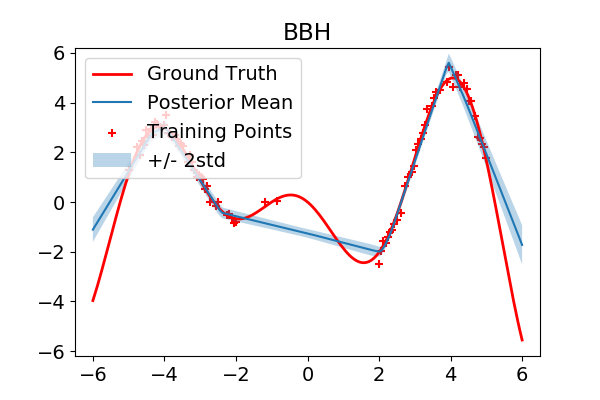} 
   ~ 
         \includegraphics[width=0.18\textwidth]{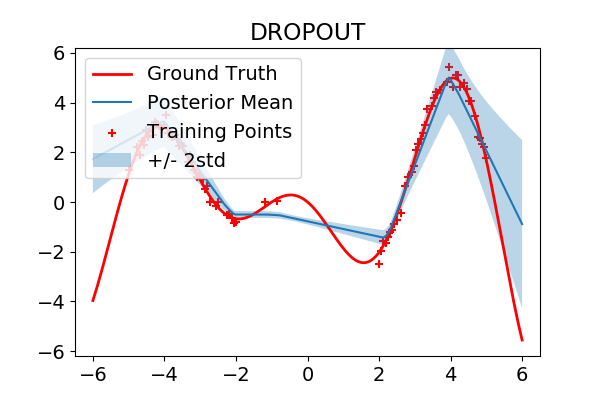}
    ~ 
         \includegraphics[width=0.18\textwidth]{toy_reg2_ensemble.png}
    ~ 
         \includegraphics[width=0.18\textwidth]{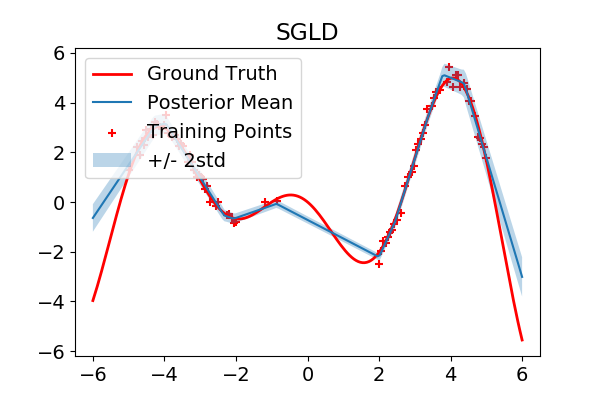}
         ~ 
         \includegraphics[width=0.18\textwidth]{toy_reg2_sghmc.png}

    \caption{A comparison of the posterior predictives for Regression with Matched A Priori Uncertainty. Ground truth (HMC) indicate that the model is calibrated. With the exception of BB-Alpha, all inference methods produce posterior predictives that model the data well over regions well represented in the training data. Of these methods, all, except for Ensemble and SGHMC, underestimate uncertainty over regions sparsely represented in the training data. }
    \label{fig:reg2_complete}
\end{figure*}

\begin{figure*}[h!]
    \centering
        \includegraphics[width=0.48\textwidth]{toy_class1_hmc.png} 
    ~
         \includegraphics[width=0.48\textwidth]{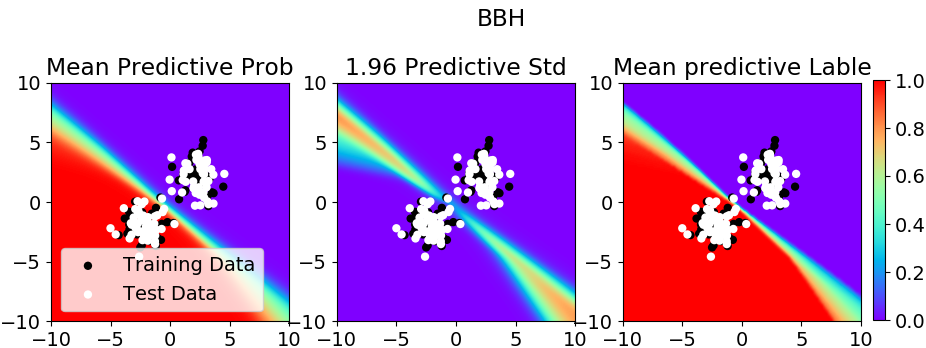} \vskip0.2cm
   ~ 
        \includegraphics[width=0.48\textwidth]{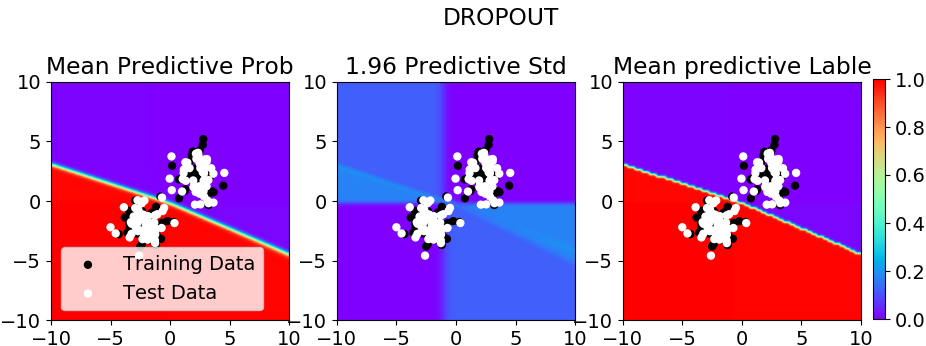} 
    ~ 
        \includegraphics[width=0.48\textwidth]{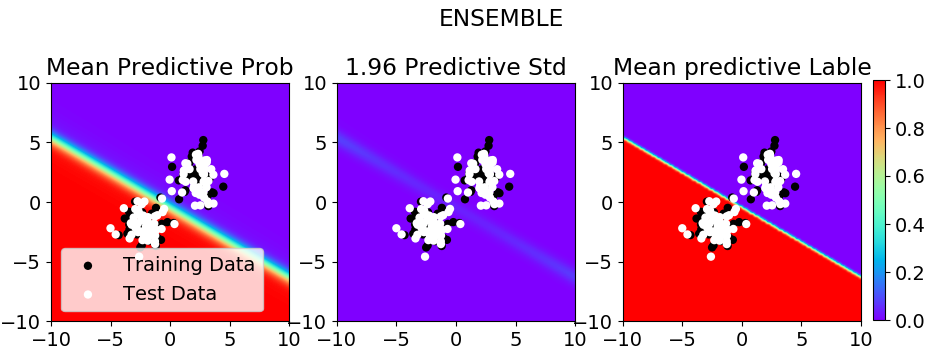} \vskip0.2cm
    ~ 
         \includegraphics[width=0.48\textwidth]{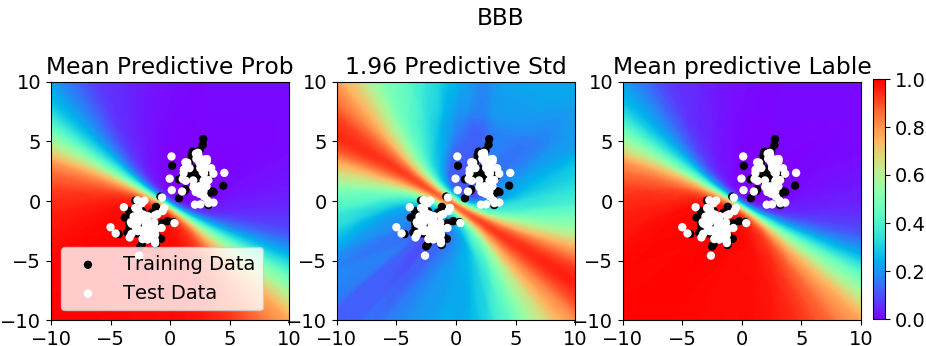} 
   ~ 
         \includegraphics[width=0.48\textwidth]{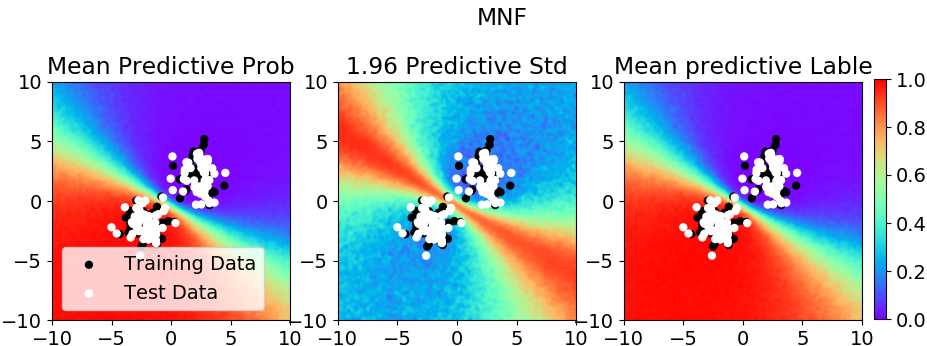} \vskip0.2cm
    ~
         \includegraphics[width=0.48\textwidth]{toy_class1_mvg.png} 
   ~ 
         \includegraphics[width=0.48\textwidth]{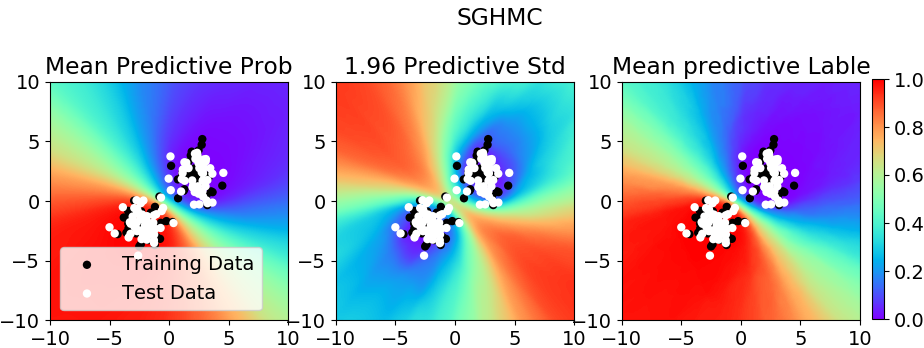} \vskip0.2cm
    ~ 
         \includegraphics[width=0.48\textwidth]{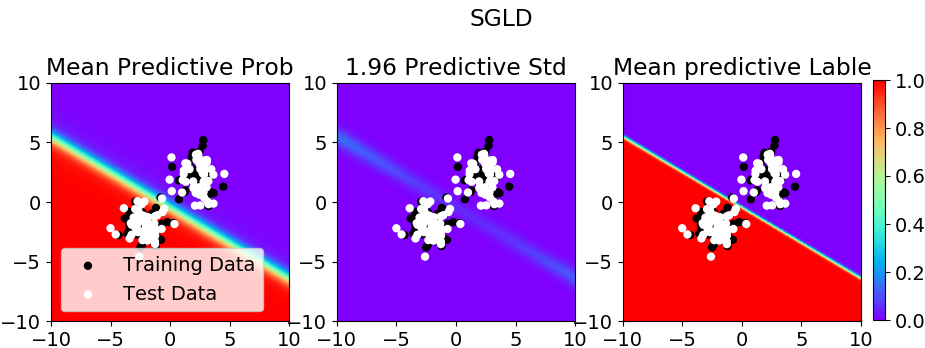}

    \caption{A comparison of the posterior predictives for Classification with Model Mismatch. Posterior predictive mean over probabilities, posterior predictive standard deviation and posterior predctive mean over labels (from left to right). Ground truth (HMC) indicates that the model is a mismatch for the data. All methods, with the exception of SGHMC underestimates predictive uncertainty.}
    \label{fig:class1}
\end{figure*}

\begin{figure*}[h!]
    \centering
        \includegraphics[width=0.48\textwidth]{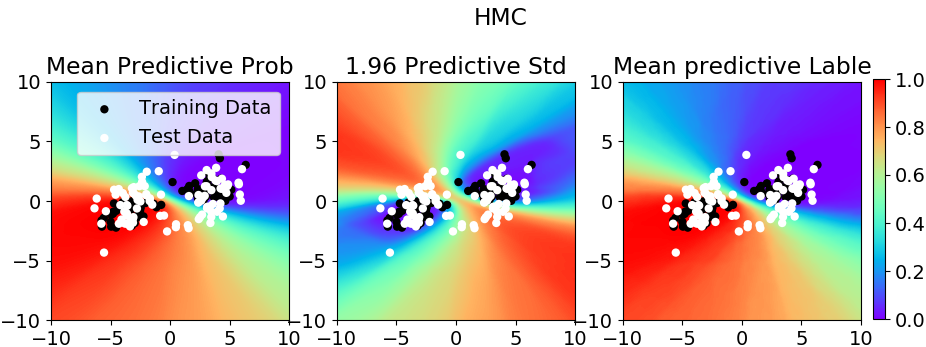} 
    ~
         \includegraphics[width=0.48\textwidth]{toy_class1_bbh.png} \vskip0.2cm
   ~ 
        \includegraphics[width=0.48\textwidth]{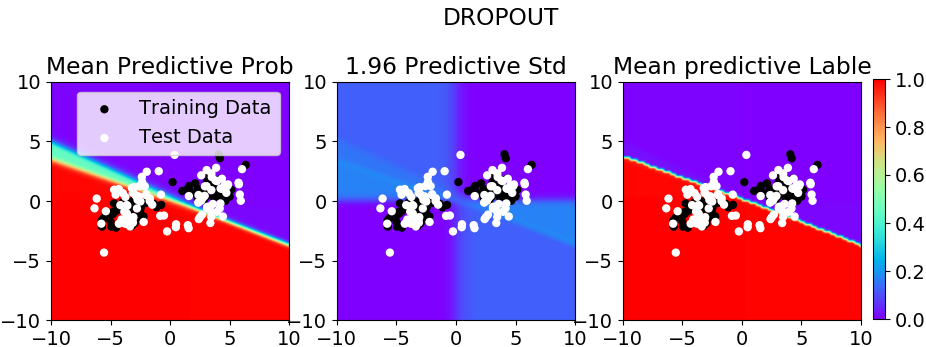} 
    ~ 
        \includegraphics[width=0.48\textwidth]{toy_class1_ensemble.png} \vskip0.2cm
    ~ 
         \includegraphics[width=0.48\textwidth]{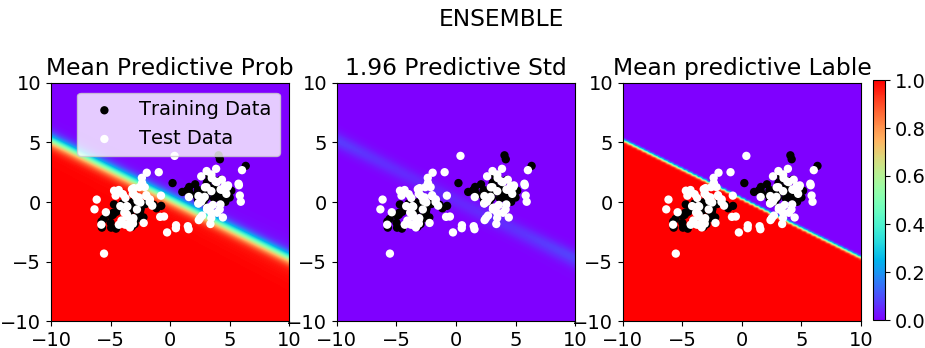} 
   ~ 
         \includegraphics[width=0.48\textwidth]{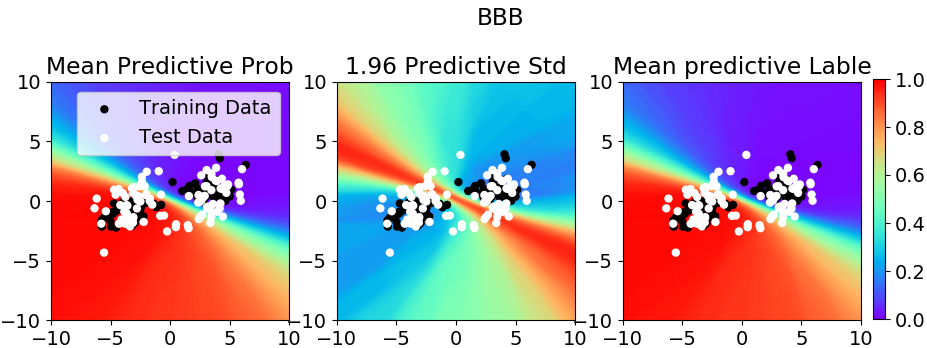} \vskip0.2cm
    ~
         \includegraphics[width=0.48\textwidth]{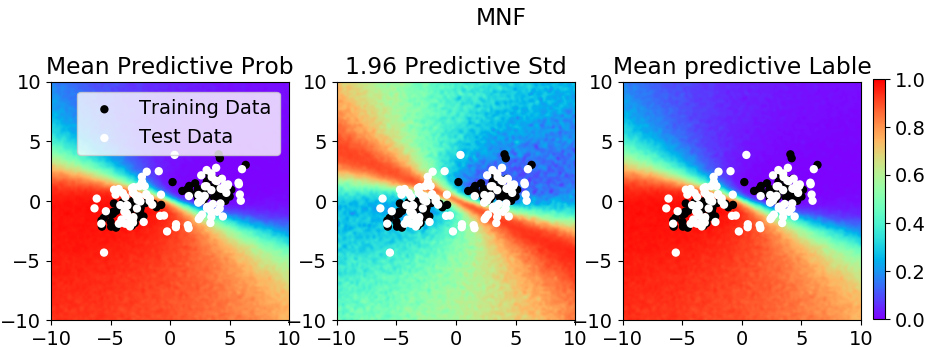} 
   ~ 
         \includegraphics[width=0.48\textwidth]{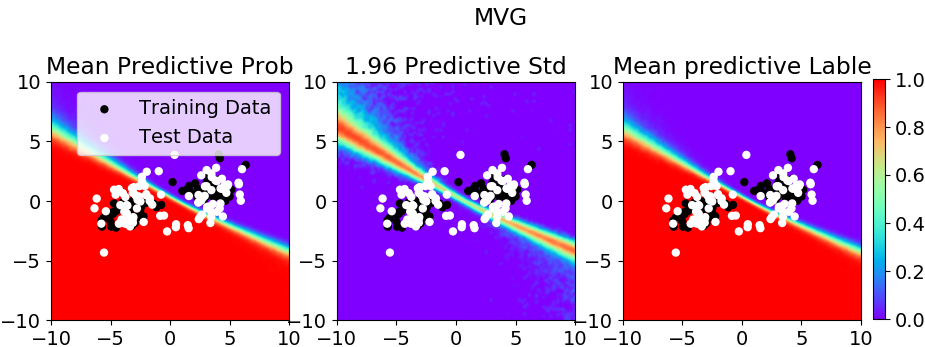} \vskip0.2cm
    ~ 
         \includegraphics[width=0.48\textwidth]{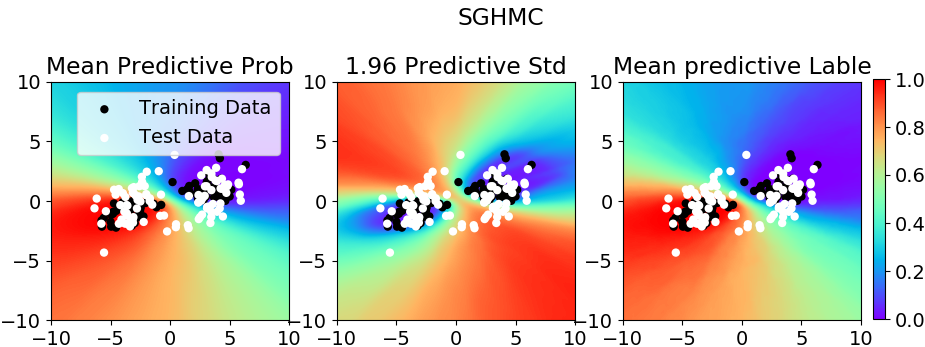}
     ~ 
         \includegraphics[width=0.48\textwidth]{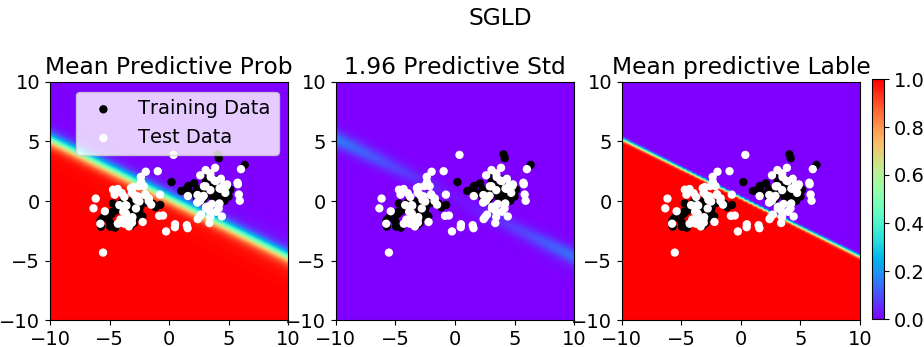}

    \caption{A comparison of the posterior predictives for Classification with No Model Mismatch. Posterior predictive mean over probabilities, posterior predictive standard deviation and posterior predctive mean over labels (from left to right). Ground truth (HMC) indicates that the model is a good match for the data. All methods, with the exception of SGHMC, underestimates predictive uncertainty. Although all methods, with the exception of SGHMC, underestimates predictive uncertainty, the effect is less severe than in the case of Classification with Model Mismatch, Figure \ref{fig:class1}.}
    \label{fig:class2}
\end{figure*}

\begin{figure*}[h!]
    \centering
     \includegraphics[width=0.18\textwidth]{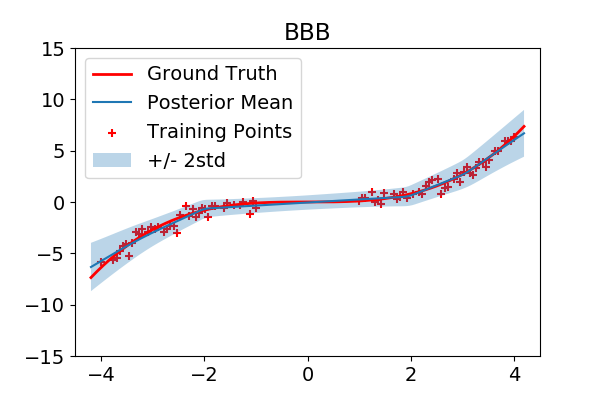} 
    ~
        \includegraphics[width=0.18\textwidth]{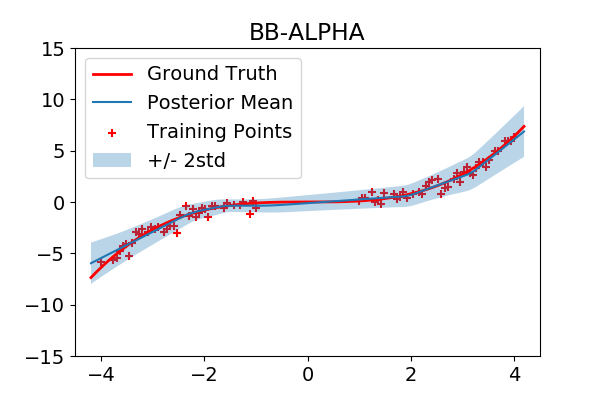} 
    ~ 
         \includegraphics[width=0.18\textwidth]{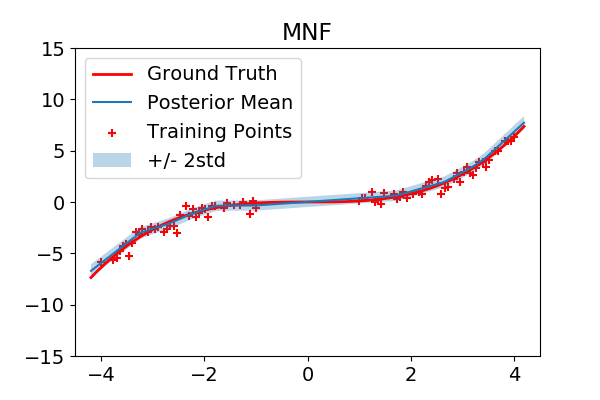} 
   ~ 
         \includegraphics[width=0.18\textwidth]{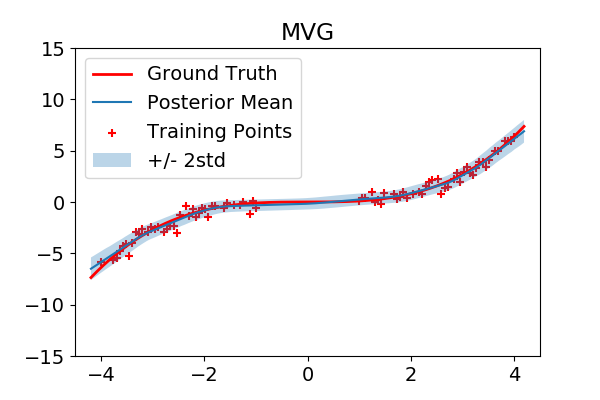}
    ~
         \includegraphics[width=0.18\textwidth]{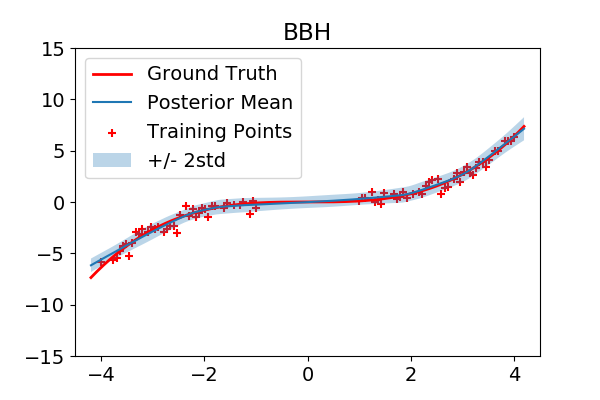} 

    \caption{The posterior predictives of models with best objective function values for Regression with Mimatched A Priori Uncertainty. }
    \label{fig:elbo_reg1}
\end{figure*}

\begin{figure*}[h!]
    \centering ~
         \includegraphics[width=0.18\textwidth]{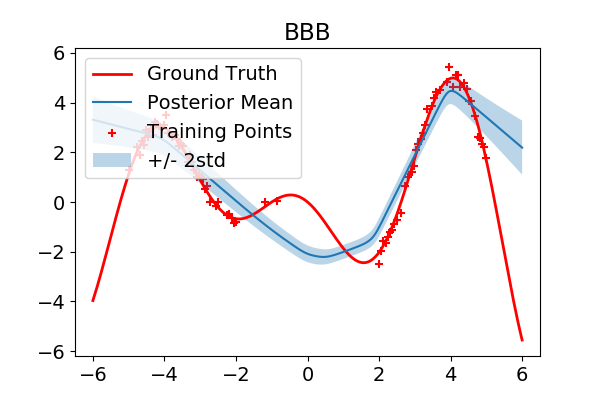} 
    ~
        \includegraphics[width=0.18\textwidth]{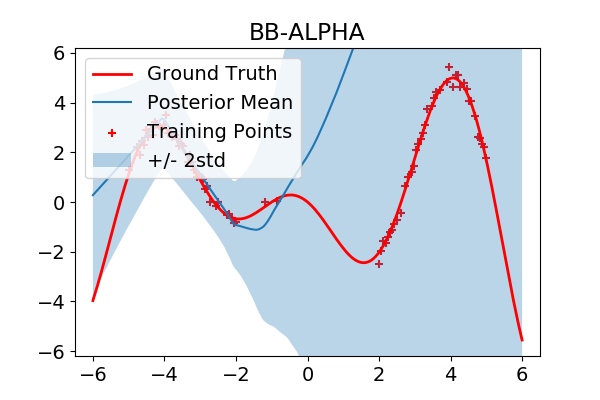} 
    ~ 
         \includegraphics[width=0.18\textwidth]{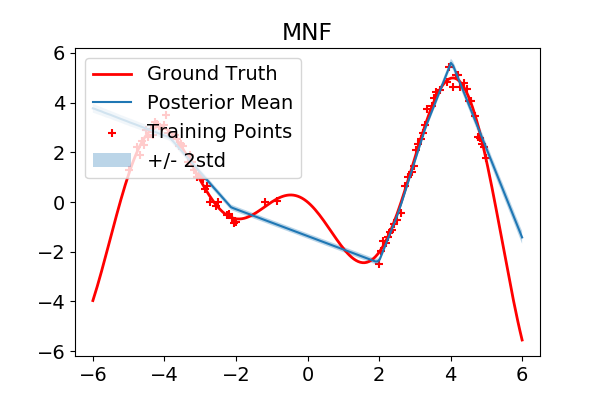} 
   ~ 
         \includegraphics[width=0.18\textwidth]{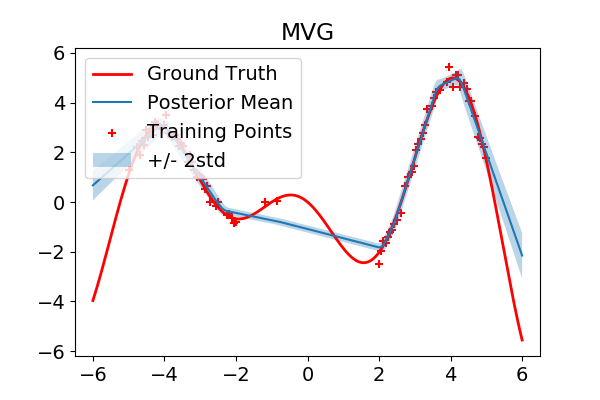}
    ~
         \includegraphics[width=0.18\textwidth]{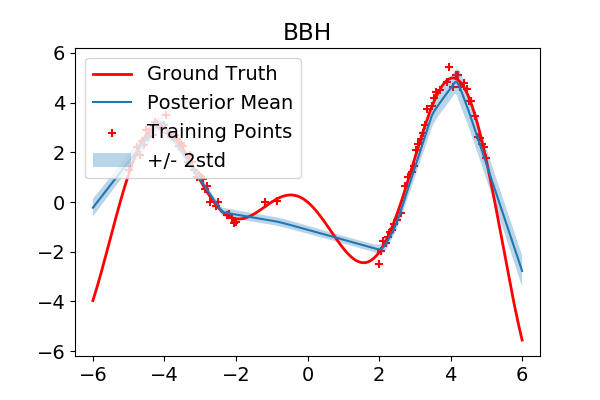} 

    \caption{The posterior predictives of models with best objective function values for Regression with Matched A Priori Uncertainty. }
    \label{fig:elbo_reg2}
\end{figure*}

\begin{figure*}[h!]
    \centering
        \includegraphics[width=0.18\textwidth]{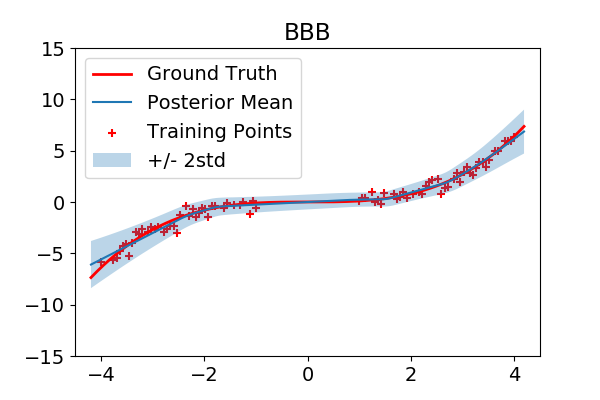} 
    ~ 
         \includegraphics[width=0.18\textwidth]{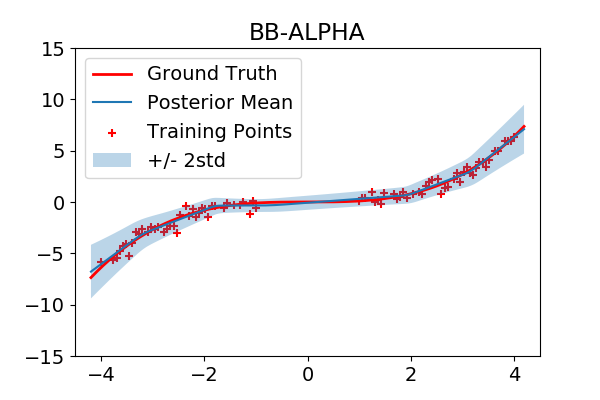} 
   ~ 
         \includegraphics[width=0.18\textwidth]{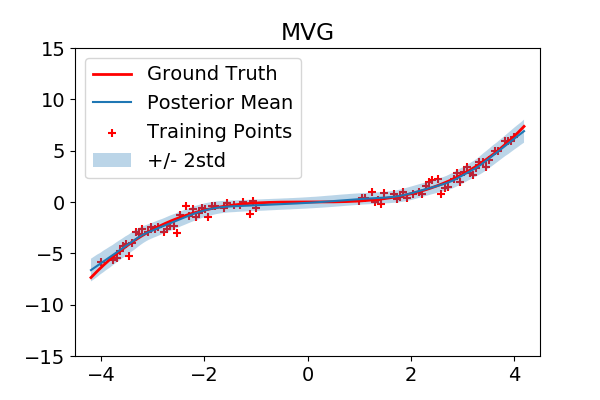}

    \caption{The posterior predictives of models of which variational parameters are initialized from the empirical mean of HMC samples for Regression with Mimatched A Priori Uncertainty. }
    \label{fig:init_reg1}
\end{figure*}

\begin{figure*}[h!]
    \centering
        \includegraphics[width=0.18\textwidth]{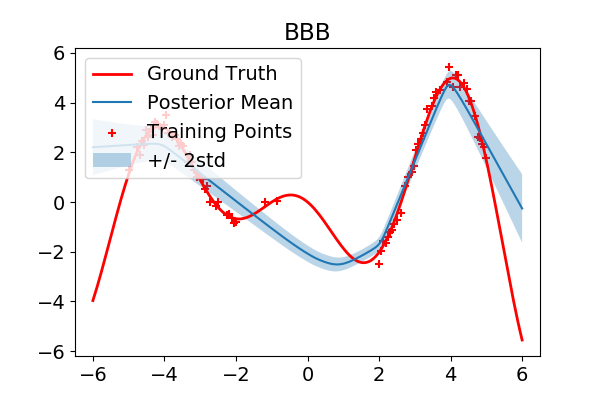} 
    ~ 
         \includegraphics[width=0.18\textwidth]{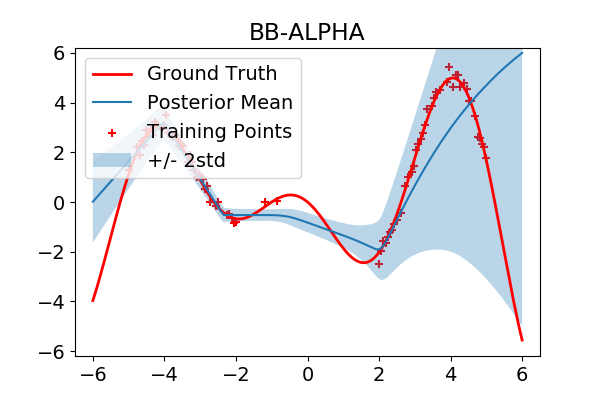} 
   ~ 
         \includegraphics[width=0.18\textwidth]{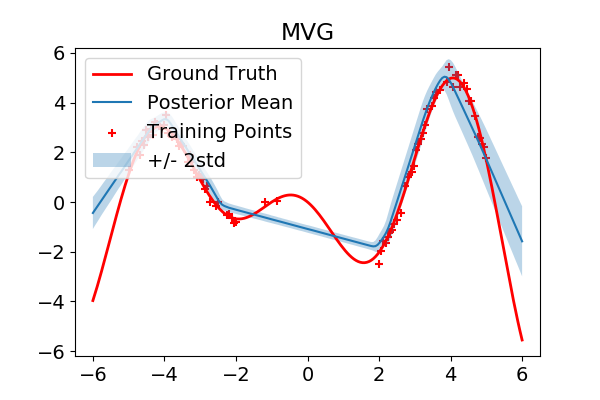} 

    \caption{The posterior predictives of models of which variational parameters are initialized from the empirical mean of HMC samples with Matched A Priori Uncertainty. }
    \label{fig:init_reg2}
\end{figure*}

\begin{itemize}
\item Figure ~\ref{fig:reg1} represents the posterior predictive distribution and Table ~\ref{tab:reg1} summarizes the metrics for all inference methods of regression task 1.
\item Figure ~\ref{fig:reg2_complete} is the complete plot of posterior predictive distribution for all inference methods of regression task 2.
\item Figure ~\ref{fig:class1} summarizes the posterior predictive distribution and Table ~\ref{tab:class1} shows the metrics for all inference methods of classification task 1.
\item Figure ~\ref{fig:class2} summarizes the posterior predictive distribution and Table ~\ref{tab:class2} shows the metrics for all inference methods of classification task 2.
\item For BBB, MNF, MVG, BBH, we ran additional experiments by using the KL divergence as model selection criterion instead of log-likelyhood. Smaller KL divergence suggests that the approximated posterior is more similar to the true posterior. For BB-ALPHA, the measurements are not comparable when $\alpha$ is different. We fixed $\alpha$ to be 0.3 (chosen by cross validation based on test loglikehood) and selected the run with the smallest $\alpha$-divergence. Figure ~\ref{fig:elbo_reg1} and Figure ~\ref{fig:elbo_reg2} summarizes the posterior predictive distribution for those models on regression tasks. Overall, those methods still do not produce satisfying approximations of the true posterior. Also, BB-ALPH does not fit the data well and significantly overestimates the uncertainty.
\item For BBB, BB-ALPHA, MVG, we helped with the optimization by initializing the variational parameters with the empirical mean of HMC samples. Figure ~\ref{fig:init_reg1} and Figure ~\ref{fig:init_reg2} summarizes the posterior predictive distribution for those models on regression tasks. Overall, the results are similar to Figure ~\ref{fig:reg1} and Figure ~\ref{fig:reg2}.
\end{itemize}

\setlength{\tabcolsep}{4.5pt}
\begin{table*}[h!]
\scriptsize
\centering
\begin{tabular}{c||c|c|c|c|c|c|c|c|c|c|c}
&\emph{HMC}&\emph{BBB}&\emph{PBP}&\emph{BB-$\alpha$}&\emph{MVG}&\emph{MNF}&\emph{BbH}&\emph{Dropout}&\emph{Ensemble}&\emph{SGLD}&\emph{SGHMC}\\
\hline
\hline
RMSE&0.18$\pm$0.01&0.26$\pm$0.02&\textbf{0.12}$\pm$0.01&0.29$\pm$0.06&0.20$\pm$0.01&0.18$\pm$0.03&0.21$\pm$0.04&0.22$\pm$0.06&\textbf{0.14}$\pm$0.00&0.16$\pm$0.01&0.22$\pm$0.01\\ 
\hline
LogLL&-0.42$\pm$0.00&-0.45$\pm$0.01&\textbf{-0.25}$\pm$0.00&-0.48$\pm$0.04&-0.34$\pm$0.01&-0.30$\pm$0.02&-0.34$\pm$0.08&-0.33$\pm$0.06&\textbf{-0.27}$\pm$0.00&-0.28$\pm$0.01&-0.53$\pm$0.01\\ 
\hline
PICP&1.00$\pm$0.00&1.00$\pm$0.00&1.00$\pm$0.00&1.00$\pm$0.00&1.00$\pm$0.00&1.00$\pm$0.00&1.00$\pm$0.00&1.00$\pm$0.00&1.00$\pm$0.00&1.00$\pm$0.00&1.00$\pm$0.00\\ 
\hline
MWPI&3.09$\pm$0.02&3.24$\pm$0.07&2.01$\pm$0.00&3.21$\pm$0.17&2.45$\pm$0.02&2.04$\pm$0.01&2.31$\pm$0.26&2.12$\pm$0.06&2.12$\pm$0.01&2.06$\pm$0.00&3.71$\pm$0.03\\ 
\end{tabular}

 \caption{A comparison of posterior predictive generalization and calibration for Regression with Mismatched A Priori Uncertainty. All methods, except for SGHMC, test log-likelihoods that are comparable if not higher to that of the ground truth (HMC). All methods, except for BBB and BB-ALPHA, have RMSE that are comparable to or lower than the ground truth. With the exceptions of BBB, BB-ALPHA and SGHMC, all methods have comparable calibration scores.}
 \label{tab:reg1}
\end{table*}
\normalsize

\setlength{\tabcolsep}{4.5pt}
\begin{table*}[h!]
\scriptsize
\centering
\begin{tabular}{c||c|c|c|c|c|c|c|c|c}
&\emph{HMC}&\emph{BBB}&\emph{MVG}&\emph{MNF}&\emph{BbH}&\emph{Dropout}&\emph{Ensemble}&\emph{SGLD}&\emph{SGHMC}\\
\hline
\hline
Accuracy&1.00$\pm$0.00&1.00$\pm$0.00&1.00$\pm$0.00&1.00$\pm$0.00&1.00$\pm$0.00&1.00$\pm$0.00&1.00$\pm$0.00&1.00$\pm$0.00&1.00$\pm$0.00\\ 
\hline
LogLL&-0.02$\pm$0.00&-0.02$\pm$0.00&\textbf{-0.00}$\pm$0.00&-0.03$\pm$0.00&\textbf{-0.00}$\pm$0.00&\textbf{-0.00}$\pm$0.00&-0.01$\pm$0.00&-0.01$\pm$0.00&-0.03$\pm$0.00\\ 
\hline
AUC&1.00$\pm$0.00&1.00$\pm$0.00&1.00$\pm$0.00&1.00$\pm$0.00&1.00$\pm$0.00&1.00$\pm$0.00&1.00$\pm$0.00&1.00$\pm$0.00&1.00$\pm$0.00\\ 
\end{tabular}

 \caption{A comparison of posterior predictive generalization for Classification with Model Mismatch. All methods, except for SGHMC, test log-likelihoods that are comparable if not higher to that of the ground truth (HMC). All methods, except for MNF and SGHMC, have log-likelihoods comparable to or higher than ground truth (HMC). All models have the same accuracy and AUC.}
 \label{tab:class1}
\end{table*}
\normalsize

\setlength{\tabcolsep}{4.5pt}
\begin{table*}[h!]
\scriptsize
\centering
\begin{tabular}{c||c|c|c|c|c|c|c|c|c}
&\emph{HMC}&\emph{BBB}&\emph{MVG}&\emph{MNF}&\emph{BbH}&\emph{Dropout}&\emph{Ensemble}&\emph{SGLD}&\emph{SGHMC}\\
\hline
\hline
Accuracy&0.83$\pm$0.01&0.82$\pm$0.01&\textbf{0.86}$\pm$0.03&0.84$\pm$0.00&0.83$\pm$0.02&0.81$\pm$0.02&\textbf{0.85$\pm$0.00}&0.84$\pm$0.00&0.83$\pm$0.02\\ 
\hline
LogLL&\textbf{-0.34}$\pm$0.01&-0.43$\pm$0.02&-2.36$\pm$0.92&-0.43$\pm$0.00&-1.12$\pm$0.14&-0.85$\pm$0.14&-0.60$\pm$0.00&-0.58$\pm$0.00&\textbf{-0.33}$\pm$0.02\\ 
\hline
AUC&\textbf{0.93}$\pm$0.00&0.92$\pm$0.01&0.91$\pm$0.03&0.91$\pm$0.00&0.89$\pm$0.01&0.89$\pm$0.02&0.90$\pm$0.00&0.90$\pm$0.00&\textbf{0.93}$\pm$0.01\\ 
\end{tabular}
 \caption{A comparison of posterior predictive generalization for Classification with No Model Mismatch. While all methods have comparable accuracy, only SGHMC has log-likelihood and AUC that as high as the ground truth (HMC).}
 \label{tab:class2}
\end{table*}
\normalsize

\section{Exploration of Structure in the HMC Posterior} 
\label{sec:posterior}
\begin{figure*}[h!]
    \centering
        \includegraphics[width=0.48\textwidth]{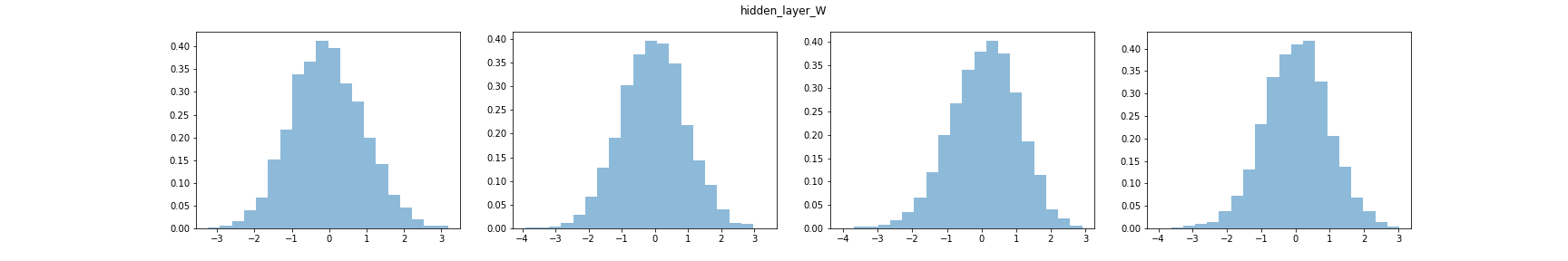} 
    ~ 
         \includegraphics[width=0.48\textwidth]{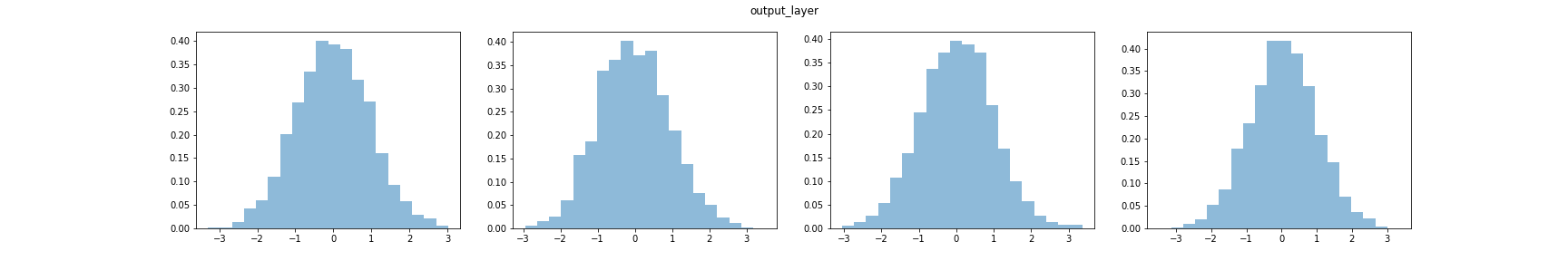} 

    \caption{Histograms of marginal posterior distributions of HMC samples.}
    \label{fig:hmc_reg1}
\end{figure*}

\begin{figure*}[h!]
    \centering
    \includegraphics[width=0.18\textwidth]{toy_reg1_hmc.png} 
    ~
        \includegraphics[width=0.18\textwidth]{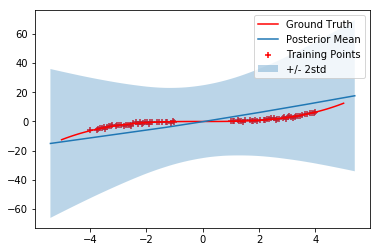} 
    \caption{The posterior predictives by approximating HMC posterior samples with a multivariate normal distribution. }
    \label{fig:appx_reg1}
\end{figure*}

We investigated the structure of the HMC posterior as it is critical to understand the types of dependencies among the weights. We found out that for regression task 1, the marginal distribution of HMC samples is close to a normal distribution, which is suggested by Figure ~\ref{fig:hmc_reg1}. We thus approximated the posterior with a multivariate Gaussian distribution $$q\sim\mathcal{N}(\mu, \Sigma)$$ where $\mu,\ \Sigma$ is approximated with
$$\hat{\mu}=\frac{1}{S}\sum_{i=1}^S w_i,\quad\hat{\Sigma}=\frac{1}{S-1}\sum_{s=1}^S (w_i-\hat{\mu})(w_i-\hat{\mu})^\intercal$$respectively and $w_i$ denotes the $i\-$th HMC sample. Figure ~\ref{fig:appx_reg1} shows that such an approximation is not sufficient to capture the dependencies among weights as both posterior mean and posterior variance is very different from ground truth. The experiment suggests that there may be higher moments correlation among the weight space. In the future, we are dedicated to investigate what types of dependencies exist in the true weight space, what types of dependencies different variational methods aim to capture, and whether those two dependencies match.

\section{Hyperparameter Settings} \label{hyperparameter}
\begin{itemize}
\item\textbf{Hamiltonian Monte Carlo (HMC)}: We implement HMC ourselves. The momentum variable is sampled from $\mathcal{N}(0,\mathit{I})$. $L=100$ leapfrog steps are used, with initial stepsize $\epsilon$ of $2\times10^{-3}$. Acceptance rate $\alpha$ is checked every $100$ iteration. $\epsilon$ is increased by $1.1$ times if $\alpha>0.8$ or decreased by $0.9$ times if $\alpha<0.2$. We used $50K$ iterations and a burnin of $40K$ and a thinning of interval 20. Convergence is verified through trace-plots and autocorrelation for weights.
\item\textbf{Bayes By Backprop (BBB)}: We tested learning rate $\epsilon\in\{0.001,0.005,0.01,0.05,0.1\}$. The code is adapted from \url{https://github.com/HIPS/autograd/blob/master/examples/bayesian_neural_net.py} 
\begin{table}[H]
\centering
\begin{tabular}{c||c|c|c|c}
&\emph{Reg 1}&\emph{Reg 2}&\emph{Class 1}&\emph{Class 2}\\
\hline
\hline
$\epsilon$&0.001&0.001&0.01&0.001
\end{tabular}
 \caption{Optimal hyperparameter for BBB.}
 \label{tab:bbb}
\end{table}
\item\textbf{Probabilistic BackPropagation (PBP)}: There are no hyperparameters to tune. We randomized the order of the data before each data sweep. The code is adapted from \url{https://github.com/HIPS/Probabilistic-Backpropagation}.  
\item\textbf{Black Box $\alpha$-Divergence (BB-$\alpha$)}:We tested learning rate $\epsilon\in\{0.001,0.005,0.01,0.05,0.1\}$ and order of $\alpha$-Divergence $\alpha\in\{0.3,0.5,0.7,1.0\}$. Note BB-$\alpha$ reduced to BBB when $\alpha=0$. The code is adapted from \url{https://bitbucket.org/jmh233/code_black_box_alpha_icml_2016}. 
\begin{table}[H]
\centering
\begin{tabular}{c||c|c|c|c}
&\emph{Reg 1}&\emph{Reg 2}&\emph{Class 1}&\emph{Class 2}\\
\hline
\hline
$\epsilon$&0.05&0.05&-&-\\
\hline
$\alpha$&0.3&0.3&-&-\\
\end{tabular}
 \caption{Optimal hyperparameter for BB-$\alpha$.}
 \label{tab:bb-alpha}
\end{table}
\item\textbf{Matrix Variate Posteriors (MVG)}: We tested learning rate $\epsilon\in\{0.001,0.005,0.01,0.05,0.1\}$. The code is adapted from \url{https://github.com/AMLab-Amsterdam/SEVDL_MGP}.
\begin{table}[H]
\centering
\begin{tabular}{c||c|c|c|c}
&\emph{Reg 1}&\emph{Reg 2}&\emph{Class 1}&\emph{Class 2}\\
\hline
\hline
$\epsilon$&0.001&0.001&0.1&0.01
\end{tabular}
 \caption{Optimal hyperparameter for MVG.}
 \label{tab:mvg}
\end{table}
\item\textbf{Multiplicative Normalizing Flows (MNF)}:We tested learning rate $\epsilon\in\{0.0005,0.001,0.005,0.01,0.0.05\}$ and the length of flow $L\in\{20,50\}$. The code is adapted from \url{https://github.com/AMLab-Amsterdam/MNF_VBNN}. 
\begin{table}[H]
\centering
\begin{tabular}{c||c|c|c|c}
&\emph{Reg 1}&\emph{Reg 2}&\emph{Class 1}&\emph{Class 2}\\
\hline
\hline
$\epsilon$&0.01&0.01&0.001&0.001\\
\hline
$L$&20&20&20&20
\end{tabular}
 \caption{Optimal hyperparameter for MNF.}
 \label{tab:mnf}
\end{table}
\item\textbf{Bayes by Hypernet (BBH)}: For regression tasks, we tested learning rate $\epsilon\in\{0.0005,0.001,0.005,0.01,0.0.05\}$ and the hypernetworks architecture $H\in\{[50],[64,64],[128,128]\}$ with nonlinearity ReLU. For classification tasks, we tested learning rate $\epsilon\in\{0.0001,0.0003,0.0005\}$ and the hypernetworks architecture $H\in\{[10],[16,16]\}$ with nonlinearity ReLU. For both tasks, five weights $w_q$ are sampled from the approximate posterior to estimate the KL divergence. The code is adapted from \url{https://github.com/pawni/BayesByHypernet/}
\begin{table}[H]
\centering
\begin{tabular}{c||c|c|c|c}
&\emph{Reg 1}&\emph{Reg 2}&\emph{Class 1}&\emph{Class 2}\\
\hline
\hline
$\epsilon$&0.001&0.001&0.0005&0.0005\\
\hline
$H$&[50]&[64,64]&[10]&[10]
\end{tabular}
 \caption{Optimal hyperparameter for BBH.}
 \label{tab:bbh}
\end{table}
\item\textbf{Dropout}: We implement Dropout ourselves, which is essentially identical to the code provided in \url{https://github.com/yaringal/DropoutUncertaintyExps}. We tested learning rate $\epsilon\in\{0.001,0.005,0.01,0.05,0.1\}$ and Bernoulli dropout rate $\gamma\in\{0.005,0.01,0.05\}$. For regression tasks, the regularization term $\lambda$ is set as the noise of the corresponding task. For classification tasks, $\lambda=0.5$. 
\begin{table}[H]
\centering
\begin{tabular}{c||c|c|c|c}
&\emph{Reg 1}&\emph{Reg 2}&\emph{Class 1}&\emph{Class 2}\\
\hline
\hline
$\epsilon$&0.05&0.05&0.005&0.01\\
\hline
$\gamma$&0.005&0.01&0.005&0.005
\end{tabular}
 \caption{Optimal hyperparameter for Dropout.}
 \label{tab:dropout}
\end{table}
\item\textbf{Ensemble}: We implement Ensemble ourselves. We tested learning rate $\epsilon\in\{0.001,0.005,0.01,0.05,0.1\}$. For regression tasks, the regularization term $\lambda$ is set as the noise of the corresponding task. For classification tasks, $\lambda=0.5$. The regularization term is chosen so that minimizing the objective function corresponds to maximizing the posterior. We collected $500$ prediction samples from $500$ random restarts.
\begin{table}[H]
\centering
\begin{tabular}{c||c|c|c|c}
&\emph{Reg 1}&\emph{Reg 2}&\emph{Class 1}&\emph{Class 2}\\
\hline
\hline
$\epsilon$&0.05&0.005&0.1&0.1
\end{tabular}
 \caption{Optimal hyperparameter for Ensemble.}
 \label{tab:ensemble}
\end{table}
\item\textbf{Stochatic Gradient Langevin Dynamics (SGLD)}: We implement SGLD ourselves. We set the batch size to be 32. We tested learning rate $\epsilon\in\{0.0005,0.001,0.005,0.01\}$. We used $500K$ iterations and a burnin of $450K$ and a thinning of interval 100.
\begin{table}[H]
\centering
\begin{tabular}{c||c|c|c|c}
&\emph{Reg 1}&\emph{Reg 2}&\emph{Class 1}&\emph{Class 2}\\
\hline
\hline
$\epsilon$&0.001&0.001&0.01&0.01
\end{tabular}
 \caption{Optimal hyperparameter for SGLD.}
 \label{tab:sgld}
\end{table}
\item\textbf{Stochatic Gradient Hamiltonian Monte Carlo (SGHMC)}: We implement SGHMC ourselves. We set the batch size to be 32. The momentum variable is sampled from $\mathcal{N}(0,\mathit{I})$. $L=100$ leapfrog steps are used and we tested stepsize $\epsilon\in\{0.001,0.002,0.005\}$. We used stepsize $\epsilon=0.002$ for all tasks. We used the friction term $C=10I$ and $\hat{B}=0$. We used $50K$ iterations and a burnin of $40K$ and a thinning of interval 20. 
\end{itemize}
\end{document}